\documentclass{article}
\usepackage[preprint]{neurips_2025}
\PassOptionsToPackage{numbers}{natbib}
\usepackage[numbers]{natbib}
\usepackage{amsmath, amssymb, amsthm}
\usepackage[utf8]{inputenc} 
\usepackage[T1]{fontenc}    
\usepackage{hyperref}       
\usepackage{url}            
\usepackage{booktabs}       
\usepackage{tabularx}
\usepackage{colortbl} 
\usepackage{amsfonts}       
\usepackage{nicefrac}       
\usepackage{microtype}      
\usepackage[dvipsnames]{xcolor}
\usepackage{graphicx}
\usepackage{xhfill}
\usepackage{placeins}
\usepackage{wrapfig}
\usepackage{float}
\usepackage{lipsum}
\usepackage{enumitem} 
\usepackage{adjustbox}
\usepackage[font=footnotesize,skip=2pt]{caption} 
\usepackage{titlesec}
\titlespacing*{\section}{0pt}{3pt}{2pt}
\titlespacing*{\subsection}{0pt}{2pt}{1pt}

\newcommand{\wrapfill}{\par\ifnum\value{WF@wrappedlines}>0
  \addtocounter{WF@wrappedlines}{-1}%
  \null\vspace{\arabic{WF@wrappedlines}\baselineskip}%
  \WFclear
\fi}
\usepackage[preprint]{neurips_2025}
\usepackage{amsmath, amssymb, amsthm}
\usepackage{bbm}
\usepackage[utf8]{inputenc} 
\usepackage[T1]{fontenc}    
\usepackage{hyperref}       
\usepackage{url}            
\usepackage{booktabs}       
\usepackage{tabularx}
\usepackage{colortbl} 
\usepackage{amsfonts}       
\usepackage{nicefrac}       
\usepackage{microtype}      
\usepackage[dvipsnames]{xcolor}
\usepackage{graphicx}
\usepackage{xhfill}
\usepackage{placeins}
\usepackage{wrapfig}
\usepackage{float}
\usepackage{lipsum}
\usepackage{enumitem} 
\usepackage{adjustbox}
\usepackage[font=footnotesize,skip=2pt]{caption} 
\usepackage{titlesec}
\titlespacing*{\section}{0pt}{3pt}{2pt}
\titlespacing*{\subsection}{0pt}{2pt}{1pt}
\usepackage{tocloft}
\usepackage{pifont}      
\usepackage{array}       
\usepackage{colortbl}    
\newcommand{\cmark}{\checkmark}
\newcommand{\lightmidrule}{\arrayrulecolor[gray]{0.7}\midrule\arrayrulecolor{black}}

\title{Escaping Plato’s Cave: JAM for Aligning Independently Trained Vision and Language Models}

\author{
  Lauren Hyoseo Yoon \\
  Computation and Neural Systems \\
  California Institute of Technology \\
  Pasadena, CA \\
  \texttt{laurenhyoon@caltech.edu} \\
   \And
   Yisong Yue \\
   Computation and Mathematical Sciences \\
   California Institute of Technology \\
   Pasadena, CA \\
   \texttt{yyue@caltech.edu} \\
   \AND
   Been Kim \\
   Google DeepMind \\
   \texttt{beenkim@google.com} \\
}

\begin{document}

\maketitle

\begin{abstract}
Independently trained vision and language models inhabit disjoint representational spaces, shaped by their respective modalities, objectives, and architectures. The Platonic Representation Hypothesis (PRH) suggests these models may nonetheless converge toward a shared statistical model of reality. This raises a fundamental question: can we move beyond post-hoc detection of such alignment and explicitly optimize for it? We argue this challenge is most critical in \textbf{fine-grained contextual distinctions}—where multiple descriptions share global semantics but differ in subtle compositional details. We address this with the Joint Autoencoder Modulator (\textbf{JAM}), which aligns frozen unimodal models by jointly training modality-specific autoencoders with coordinated reconstruction and cross-modal alignment objectives. We systematically evaluate JAM across three design axes: (i) alignment objectives, introducing our multimodal Spread Loss that outperforms classic contrastive methods; (ii) the layer depth at which alignment is most effective; and (iii) the role of foundation model scale in representational convergence. Our findings show that JAM reliably induces alignment even across independently trained representations, offering both theoretical insight into the structure of shared semantics and practical guidance for transforming generalist unimodal foundations into specialist multimodal models.
\end{abstract}




\section{Introduction}

Neural networks trained on different modalities, datasets, and objectives typically inhabit disjoint representational spaces. Yet the Platonic Representation Hypothesis (PRH) \cite{platonic} suggests that these models—despite having no shared supervision, architecture, or training regime—may nonetheless converge toward a common statistical model of reality. The naming of PRH is inspired by Plato’s Allegory of the Cave \cite{plato}, which describes individuals who observe only shadows of objects cast on the wall of a cave and mistake these projections for the entirety of reality. This hypothesis has been discussed under several philosophical and empirical lenses, including convergent realism in the philosophy of science and the Anna Karenina scenario \cite{stitching} in representation learning, which suggests that all well-performing models may ultimately resemble each other. 

However, most evidence remain observational, relying on statistical metrics that infer compatibility at a \textit{coarse-grained level}. These methods quantify global correlations across feature spaces, but do not provide practical mechanisms for constructing multimodal systems from unimodal ones, nor do they illuminate where alignment fails. Crucially, the limitations emerge in \textit{fine-grained contextual settings}, where multiple candidate descriptions may share overall semantics yet differ in specific details. For example, distinguishing whether an image contains a dog is a coarse judgment, but deciding between ``a brown dog chasing a red ball’’ and ``a brown dog chasing a blue ball’’ requires contextual sensitivity to subtle compositional cues. It is precisely in these cases (i.e., where local semantic differences matter) that the true challenge of alignment lies. 

\subsection{Formulation of Plato's Cave}
We cast Plato's Cave in two regards:
\vspace{-0.07in}
\begin{itemize}[leftmargin=*, itemsep=0.5pt]
    \item \textit{Unimodal representations:} 
    Borrowing from the Plato's Cave allegory, each unimodal data and training regime yields only a ``shadowed projection’’ of reality. Our goal is to uncover the latent, Platonic representation: a shared semantic structure that exists despite the models being trained in isolation.
    \item \textit{Fine-grained context:}
    While unimodal embeddings capture coarse contextual structure (e.g., broad semantic overlap), they often fail to encode discriminative fine-grained details. Here, ``context’’ refers to the high-level semantics shared across modalities, while ``fine-grained’’ refers to resolving distinctions within that shared context (e.g., attributes, relations, or localized compositional shifts).
\end{itemize}

\subsection{Escaping Plato's Cave (Platonic Alignment)}
Motivated by these insights, we propose \textbf{Platonic Alignment}, a framework to explicitly align unimodal models trained independently on distinct modalities. Our Joint Autoencoder Modulator (\textbf{JAM}, Fig.~\ref{fig:jam}) aligns frozen language and vision representations using coordinated reconstruction and alignment objectives. Reconstruction preserves modality-specific information, while a shared bottleneck enforces a coherent conceptual space capable of resolving both coarse and fine-grained semantics. 

Our contributions are as follows:
\vspace{-0.07in}
\begin{itemize}[leftmargin=*, itemsep=0.5pt]
    \item We introduce the multimodal \textbf{Spread Loss}, which leverages contextual structure to outperform classic contrastive objectives for fine-grained alignment. 
    \item We demonstrate the versatility of our \textbf{JAM} framework across a wide range of pretrained backbones, and validate its components through ablations and comparisons to projection-based baselines. 
    \item We analyze the impact of pretrained backbone/model's scale and layer depth on alignment performance, providing insights into how these factors interact with alignment supervision.
\end{itemize}

\subsection{Related Work} 
Our work is related to two research threads: analyzing correlations between unimodal representations, and aligning different data modalities to construct multimodal models.

\textbf{Testing for Alignment.} 
Prior work in this direction has been largely diagnostic focusing on evaluating alignment between frozen features with broad, context-agnostic datasets (e.g.,  Wikipedia caption dataset (WIT) \cite{wit}): measures such as centered kernel alignment (CKA) \cite{cka}, variants of CCA (SVCCA \cite{svcca}, projection-weighted CCA \cite{morcos2018insights}), and nearest-neighbor metrics \cite{nearest_neighbors, platonic}. Other approaches explore probing tasks or zero-shot transfer to assess latent compatibility across modalities. More recently, model-stitching frameworks have examined whether independently trained sub-networks can be functionally composed to perform new tasks \cite{stitching}, hinting at deeper interoperability between pretrained systems but focusing on vision domain \cite{stitching, stitching2}.

While informative, these methods are designed to be passive—they measure alignment but do not offer mechanisms to optimize or induce it. As a result, they may conflate superficial correlation with functional compatibility: two embedding spaces may appear statistically aligned yet remain ineffective for fine-grained multimodal tasks. Moreover, although recent state-of-the-art multimodal models (e.g., Gemini \cite{gemini2023}, GPT-4V \cite{openai2023gpt4v}, LLaMA 3 \cite{llama3modelcard}) demonstrate strong cross-modal performance, they do not explicitly address the alignment dynamics between independently trained unimodal components. In contrast, our work provides a systematic and controlled framework for probing and optimizing cross-modal alignment—offering a potential design lens for future multimodal systems that build on or unify strong unimodal foundations. 

\textbf{Optimizing Alignment of Representations.} 
A common framing of alignment in multimodal learning refers to the emergence of structurally coherent or comparable latent spaces across modalities, such that semantically related inputs (e.g., images and their captions) map to nearby embeddings. This framing underlies much of the recent progress in large-scale multimodal models such as CLIP \cite{clip}, ALIGN \cite{jia2021scaling}, and BLIP/BLIP-2 \cite{li2022blip, li2023blip2}, DeepSeek \cite{deepseekvl2024}, which are  trained end-to-end using massive paired corpora using explicitly multimodal objectives. Notably, BLIP/BLIP-2 adopt a modular architecture that connects frozen vision encoders and large language models, offering design flexibility. However, the pretrained vision encoders utilized in this method is CLIP, thus inheriting CLIP’s multimodal alignment objective from the outset. In contrast, our approach begins with truly unimodal foundations—vision and language models trained independently—and investigates whether alignment can emerge post hoc, without relying on pre-imposed multimodal inductive biases.

\section{Statistical Tests for Representation Alignment}

\begin{figure}[t] \centering
    \includegraphics[width=\textwidth]{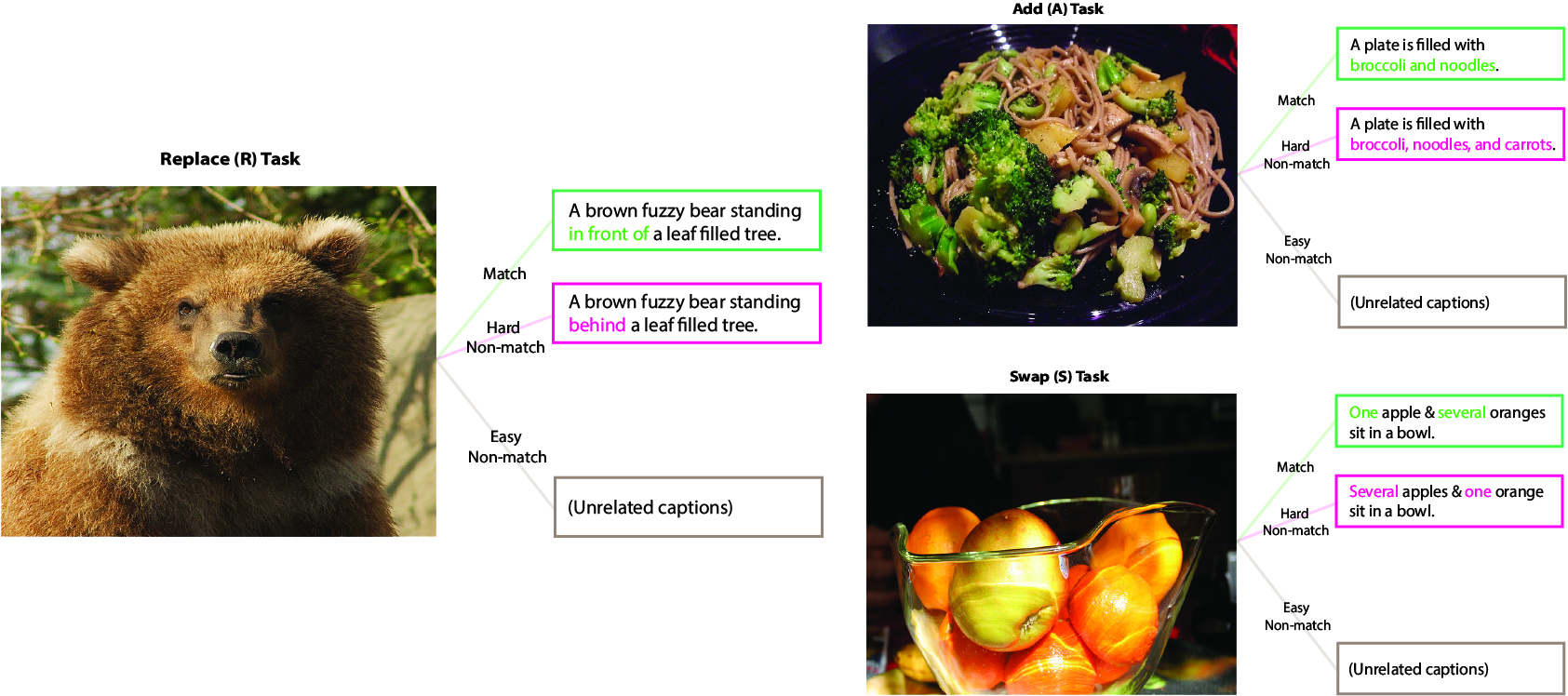}
    \vspace{-0.13in}
    \caption{\scriptsize Illustration of fine-grained contextual understanding from the SugarCrepe dataset \cite{hsieh2023sugarcrepe}. Each image is paired with three types of captions: (i) \textbf{Match} (true positive) captions that correctly describe the image, (ii) \textbf{Easy non-match} captions that are entirely unrelated, and (iii) \textbf{Hard non-match} (hard negative) captions that share global semantics with the true caption but diverge in subtle, fine-grained details (e.g., swapping relations, replacing objects, or adding attributes). These controlled perturbations—Replace (R), Swap (S), and Add (A) from the SugarCrepe dataset \cite{hsieh2023sugarcrepe} provide a principled test for whether alignment methods capture \textit{contextual fine-grained distinctions} rather than just coarse semantic similarity.}
    \label{fig:data}
    \vspace{-0.11in}
\end{figure}

To probe the potential for aligning unimodal models, we applied four alignment metrics—CCA, CKA, SVCCA, and CKNNA—to three types of image–text pairs, illustrated in Fig.~\ref{fig:data}: 
\textbf{match} (true positive, image–caption pairs), \textbf{easy non-match} (unrelated captions), and \textbf{hard non-match} (semantically similar captions differing in fine-grained details). The hard negatives are particularly vital in our setting: While they share global semantics with the true captions, they diverge in localized compositional attributes, testing whether alignment captures fine-grained context rather than coarse similarity.

We evaluate embeddings from unimodally pretrained vision and language models (Gemma2 (2B) \cite{gemma_2024}, Llama3.2 (1B)  \cite{llama3modelcard}, OLMo2 (7B) \cite{olmo20242olmo2furious} for language; DINOv2 (ViT-B) \cite{oquab2023dinov2}, ResNet50 \cite{he2016deep} for vision), using the SugarCrepe dataset \cite{hsieh2023sugarcrepe}, which is explicitly designed for fine-grained vision–language compositionality. SugarCrepe provides minimal caption perturbations across three transformation families—Replace, Swap, and Add—yielding controlled hard negatives (see Fig.~\ref{fig:data}). For experiments, we extract CLS-token image embeddings and penultimate-layer text embeddings, then compute alignment scores using the four metrics (See Appendix for metrics formulations).

\begin{figure}[t] \centering
    \includegraphics[width=\textwidth]{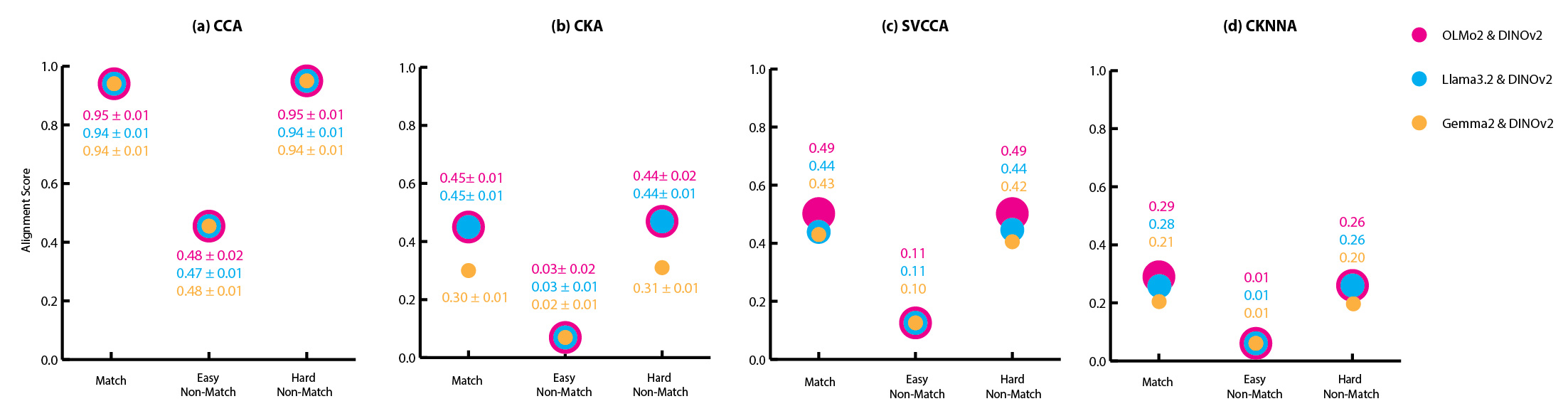}
    \caption{\scriptsize Statistical Metrics for Representation Alignment: Across all metrics and model cases, match pairs consistently show higher alignment scores than easy non-match pairs, supporting the hypothesis that unimodal models encode shared global structure. However, hard non-match pairs exhibit similarly high scores. This indicates that while statistical metrics for representations reveal coarse representational compatibility -- which aligns with PRH \cite{platonic} -- but are insufficient for diagnosing semantic alignment at a more granular level. (a) Linear and Kernel (Gaussian (i.e., RBF) kernel) PCA to reduce the embeddings dimension to 50; Set CCA dimension as 50. Variation of score values are based on the utilized kernel. (b) Variation of score values are based on the utilized kernel (Linear \& RBF). (c) SVD to reduce embeddings dimension to 10; Set CCA dimension as 10. (d) Set top k nearest neighbors = 10 (following the default setting of \cite{platonic}).}
    \label{fig:metrics}
    \vspace{-0.21in}
\end{figure}

With the extracted feature representations 
\footnote{All experiments were conducted with non-GDM (Google DeepMind) compute resources.}, let $D$ be the whole data, and $B$ be the batch. We construct $D$ in the nested pair format: $D = \{(v_i, (l_{P_i}, l_{N_i})\}_{i=1}^{n}$ (Refer to Table \ref{glossary} for variable descriptions). $C(i)$, the set of similar context text embedding for anchor $i$, is defined as $l_{P_i}$ and its hard negative $l_{N_i}$ (i.e., $C(i) = \{l_{P_i}, l_{N_i}\}$). Accordingly, $\widetilde{C}(i) = L \setminus C(i)$. (In Table \ref{fig:metrics}, we show results for DINOv2 as vision backbone. For full results on refer to Appendix Table \ref{tab:replace_stats},\ref{tab:add_stats},\ref{tab:swap_stats}.)

\subsection{Statistical tests detect alignment, but not fine-grained context.}

The results (Fig.\ref{fig:metrics}) show that all models reliably distinguished matching from easy non-matching pairs, confirming that independently trained vision and language models encode a shared high-level structure consistent with the Platonic Representation Hypothesis \cite{platonic}. However, in the hard non-match condition, statistical metrics produced scores nearly indistinguishable from true matches. This is not a failure of alignment but a reflection of the metrics themselves: hard negatives share global semantics with positives, differing only in fine-grained attributes. \textbf{We conjecture that representational similarity is real at the coarse level but insufficiently sensitive to discriminative detail.} Thus, while statistical tests reveal broad compatibility, they obscure nuanced distinctions critical for fine-grained alignment. To move beyond this limitation, we turn to a lightweight post-hoc training approach that explicitly optimizes for semantic coherence, both globally and locally. This insight directly led to our alignment training scheme in the next section.

\section{Our method: Joint Autoencoder Modulator (JAM)}

\begin{wrapfigure}{r}{0.58\textwidth} 
  \vspace{-5mm}
  \centering
  \includegraphics[width=0.56\textwidth]{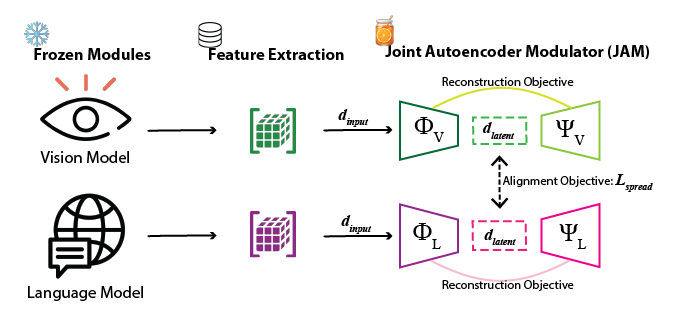}
  \caption{Joint Autoencoder Modulator (JAM) framework.}
  \label{fig:jam}
  \vspace{-3mm}
\end{wrapfigure}

Our approach learns a joint autoencoder across two disjoint pre-trained models, as shown in Figure \ref{fig:jam}. We first utilize a modality‑specific autoencoder for each pre-trained model which distills $d_{input} \rightarrow d_{latent}$, which can be viewed as a form of statistical regularisation reminiscent of PCA \cite{regae, regae_pc}: the network is encouraged to find a compact set that preserve essential variance while discarding noise through reconstruction loss (i.e., MSE Loss). 
The joint training objective then contrastively pulls the latent vectors from matched image–text pairs together and push hard‑negatives apart. Thus, the respective autoencoders have two objectives: denoising for each modality as well as optimizing for cross-modal relation (the exact loss functions below).

\begin{wraptable}{r}{8.8cm}
\vspace{-0.15in}
\scriptsize
\caption{Glossary of variables and symbols}
\label{glossary}
\centering
    \begin{tabularx}{8.8cm}{cl}
    \hline
    Symbol & Description \\ 
    \hline
    $\Phi_V, \Psi_V$ & Vision (image) autoencoder (encoder, decoder) \\
    $\Phi_L, \Psi_L$ & Language (text) autoencoder (encoder, decoder) \\
    $V$ & Set of image embeddings ($|V| = n, v_i \in V, i=\{1,\dots, n\}$) \\
    $L$ & Set of text embedding (i.e., $L_P \cup L_N$; $|L| = 2n$) \\
    $L_P$ & Set of positive text embeddings ($|L_P| = n, l_{P_i} \in L_P$) \\
    $L_N$ & Set of \textbf{hard} negative text embeddings ($|L_N| = n, l_{N_i} \in L_N$) \\ 
    $d_{input}, d_{latent}$ & Input, Latent (i.e., bottleneck) dimension of autoencoder \\ 
    $l_i$ & Element of $L \setminus \{l_{P_i}\}$ with anchor index $i$ \\ 
    $C_L(i)$ & Set of similar context text embeddings \\
    $\widetilde{C_L}(i)$ & Set of dissimilar context text embeddings (i.e., $L \setminus C_L(i))$ \\
    \hline
    \end{tabularx}
\end{wraptable}

The encoder follows a “funnel” layout: a sequence of fully connected layers that progressively reduce the dimensionality of the input embedding, each stage wrapped in LayerNorm for stable statistics \cite{ba2016layernormalization}, a SwiGLU non‑linearity for gated expressiveness \cite{shazeer2020gluvariantsimprovetransformer}, and dropout for regularisation. After every dense layer, a lightweight MLP residual block \cite{vaswani2023attentionneed} re‑injects the intermediate representation back into itself, ensuring gradient flow. Passing through these hidden layers, in a bottleneck stage, a final linear projection maps the hidden width to the fixed latent size. The decoder mirrors the encoder, walking back through the hidden sizes in reverse. Together, we chose this design to be an effective framework for building compact yet powerful module that fits both high‑dimensional text and vision representations (i.e., embeddings). In our experiments, we used 3 hidden layers with dimension size of 512, and bottleneck dimension size as 256. 

\subsection{Loss functions}

For the objective design (i.e., loss functions) to jointly train autoencoders for alignment (along with default reconstruction objective), we adopt Contrastive and Negative Contrastive loss, which are utilized as baseline methods, and further introduce our new loss scheme: Spread Loss functions for vision-to-language and language-to-vision. 

Having a set of losses designed to optimize different aspects of the alignment had a side benefit of helping us to diagnose and analyze experiments (e.g., which layer to establish alignment), as will be shown in later sections. 

\paragraph{Contrastive Loss (Con)}
Same as the contrastive loss for training CLIP model \cite{clip}, we formulate our contrastive loss ($\mathcal{L}_{con}$), in a symmetric fashion, designed to enforce bidirectional consistency between images and texts: Vision-to-Language (VL) as matching each image with its corresponding text, and Language-to-Vision (LV) as matching each text with its corresponding image. We then leverage cross-entropy to maximize the similarity between the correct pairs while minimizing the similarity between incorrect pairs. The cross-entropy loss is given by $\mathcal{L} = - \sum_{i=1}^N y_i \log p_i$, where $y_i$ is the ground truth label ($y_i = 1$ for the correct pair and $y_i = 0$ for incorrect pairs), $p_i$ is the predicted probability of the correct pair. 
Thus, the final contrastive loss is formulated as:
$\mathcal{L}_{con} = \frac{1}{2} \left[ \mathcal{L}_{\text{VL}} + \mathcal{L}_{\text{LV}} \right]$. 
where image-to-text loss (VL) and text-to-image (LV) loss are the following, respectively:  
{\footnotesize
\begin{equation}
\begin{split}
\mathcal{L}_{\text{VL}} = - \frac{1}{N} \sum_{i=1}^N \log \frac{\sigma(v_i, l_{P_i})}{\sum_{j=1, j \neq i}^N \sigma(v_i, l_{P_j})},\quad \mathcal{L}_{\text{LV}} = - \frac{1}{N} \sum_{i=1}^N \log \frac{\sigma(l_{P_i}, v_i)}{\sum_{j=1, j \neq i}^N \sigma(l_{P_i}, v_j)}
\end{split}\label{eq:Con}
\end{equation}
}\noindent
Here, we denote $\mathcal{\sigma}(x, x') = \exp(\text{sim}(x, x') / \tau)$, where $x, x'$ are the modality-specific embeddings (in this case, text and image embeddings, respectively), and $\tau$ is the temperature parameter that controls the sharpness of the similarity distribution. For all the experiments, we set $\tau=0.07$, as done in \cite{clip}. 

\paragraph{Negative Contrastive Loss (NegCon)} 
The standard CLIP objective treats all non-matching text descriptions as implicitly negative, but without distinguishing between truly unrelated captions and hard-to-distinguish negative examples (hard negatives)—those that share high semantic similarity with the positive caption but are incorrect due to fine-grained distinctions (e.g., a location or object mismatch). This lack of granularity can lead the model to underutilize structurally informative negative samples. We address this challenge by introducing NegCon loss ($L_{NegCon}$), which allows hard negative texts to also be penalized in the image-to-text direction, while keeping the text-to-image direction unchanged for stability. This loss scheme to incorporate hard negative texts is what has been used in the development of NegCLIP \cite{yuksekgonul2023when}. 

For the vision-to-language direction, we extend the candidate pool of possible text matches by concatenating both positive and hard negative captions, effectively doubling the number of candidates:  
The image-to-text loss is then defined as:
{\footnotesize
\begin{equation}
\begin{split}
\mathcal{L}_{\text{VL}} &= - \frac{1}{2N} \sum_{i=1}^N \log \frac{\sigma(v_i, l_{P_i})}{\sum_{j=1, j \neq i}^{2N} \sigma(v_i, l_{j})}
\end{split}\label{eq:NegCon}
\end{equation} 
}\noindent
This formulation explicitly anchors the image to its positive caption while contrasting it against a broader and more semantically informative set that includes hard negatives.

For the language-to-vision direction, we retain a standard CLIP-style formulation, utilizing $\mathcal{L}_{\text{LV}}$ utilized in $\mathcal{L}_{con}$. Since hard negative captions have no associated image, we only use positive captions and match them against the corresponding images. 

The final NegCon loss is: $\mathcal{L}_{NegCon} = \frac{1}{2} \left[ \mathcal{L}_{\text{VL}} + \mathcal{L}_{\text{LV}} \right]$. This symmetric structure ensures that both modalities contribute to alignment, while the asymmetry in negative usage avoids destabilizing supervision on text inputs with unknown image counterparts. Hence NegCon provides a minimally modified but effective baseline for studying the role of hard negative supervision. 

\paragraph{Spread Loss}
Recall that our objective is to develop a loss function that more faithfully captures the structure of semantic similarity and fine-grained contrast in multimodal representation learning. Specifically, we aim to align image and text embeddings in a way that accounts for both shared contextual meaning and localized distinctions (fine-grain). Standard contrastive losses treat all non-matching text as equally negative, which neglects the semantic proximity of hard negatives—text descriptions that differ from the correct caption only in fine detail.

\begin{wrapfigure}{R}{0.3\textwidth}
\vspace{-4mm}
     \includegraphics[width=0.25\textwidth]{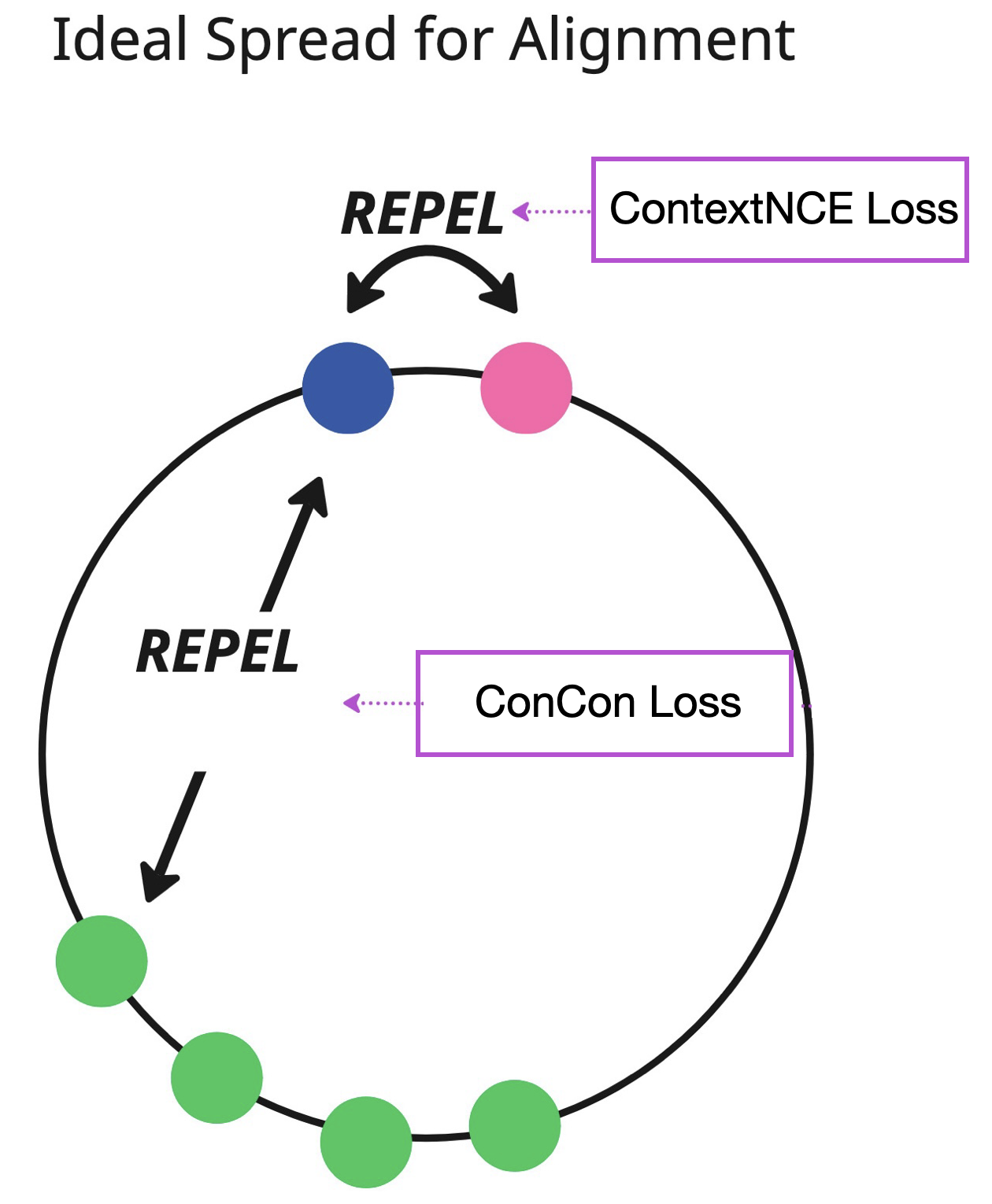}
     \caption{\scriptsize Illustration of the Objective of Spread Loss Formulation: \textcolor{NavyBlue}{Blue} and \textcolor{WildStrawberry}{Pink} circle correspond to similar context group; \textcolor{ForestGreen}{Green} circles are representations outside of similar context. Figure inspired by \cite{chen2022perfectly}. }
     \label{fig:spread}
\vspace{-0.5mm}
\end{wrapfigure}
To address this, we introduce $\mathcal{L}_{spread}$, a contrastive objective that incorporates a notion of context similarity and fine-grained differentiation. As shown in Figure \ref{fig:spread}, the core idea is to pull together all semantically similar captions, including the ground-truth caption and selected hard negatives, and then selectively push apart fine-grained distinctions within that similar group. This two-stage formulation allows us to preserve shared contextual alignment while enforcing discriminative power at the instance level. The idea of spread loss has been proposed in the domain of visual representation learning \cite{chen2022perfectly, fu2022details}, which is a variant of Supervised Contrastive (SupCon) loss \cite{supcon}, but tailored to tackling class collapse.
We take inspiration from that work and apply this idea to the multi-modal alignment problem.\footnote{We provide a parallel comparison between $\mathcal{L}_{spread}$ of multimodal representation space (i.e., ours) and $\mathcal{L}_{spread}$ of visual representation domain \cite{chen2022perfectly, fu2022details} in the Appendix.}
Our $\mathcal{L}_{spread}$ is formulated as:
{\footnotesize
\begin{equation}
\mathcal{L}_{spread} = \frac{1}{2} \left [\mathcal{L}_{spread-\text{VL}} + \mathcal{L}_{\text{LV}} \right]
\end{equation}
}\noindent
where $\mathcal{L}_{spread-\text{VL}} = (1-\alpha)\mathcal{L}_{ConCon} + \alpha \mathcal{L}_{contextNCE}$ and $\mathcal{L}_{\text{LV}}$ is from \eqref{eq:Con}, preserving bidirectional learning throughout joint-AE training.

$\mathcal{L}_{ConCon}$ : 
ConCon (i.e., Context Contrastive) lumps the positive \& hard negative (which share contextual attributes) case as the similar context set and tries to push this set away from all other text embeddings. The objective of this loss is to introduce the global learning scheme where we focus on learning what is easily differentiable. Formally, we construct the loss as follows: 
{\footnotesize
\begin{equation*}
\begin{split}
    \mathcal{L}_{ConCon}(i, B) &= - \frac{1}{|C_L(i)|}\sum_{c_l \in  C_L(i)} \log \frac{\sigma (v_i, c_l)}{\sigma(v_i, c_l) + \sum_{\widetilde{c_l} \in \widetilde{C_L}(i)} \sigma(v_i, \widetilde{c_l})}
\end{split}
\end{equation*}
}\noindent
$\mathcal{L}_{contextNCE}$ : 
We introduce this second term for local learning scheme. We perform an InfoNCE-style objective, explicitly highlighting the true positive as the only “correct” text. The negative text embedding, though labeled as 'similar context', is still in the denominator and is therefore treated as incorrect text. 
{\footnotesize
\begin{equation*}
\begin{split}
    \mathcal{L}_{contextNCE}(i, B) &= - \log \frac{\sigma(v_i, l_{P_i})}{\sum_{c_l \in C_L(i)} \sigma(v_i, c_l)}
\end{split}
\end{equation*}
}\noindent
The final formulation for VL direction is a weighted sum: 
$\mathcal{L}_{spread-\text{VL}} = (1-\alpha)\mathcal{L}_{ConCon} + \alpha\mathcal{L}_{contextNCE}$, where $\alpha$ controls the trade-off between context-level alignment and intra-context contrast. While $\mathcal{L}_{ConCon}$ allows similar context to be close, $\mathcal{L}_{contextNCE}$ prevents representation collapse in similar context embeddings. Thus, $\mathcal{L}_{spread}$ enables learning representations that are not only globally aligned across modalities but also sensitive to subtle mismatches in local content. In our experiments, we use $\alpha=0.5$ to balance the two components in $\mathcal{L}_{spread-\text{VL}}$. 


\section{Experiments}
We adopt the same data set-up as in the statistical tests using Sugarcrepe \cite{hsieh2023sugarcrepe}, splitting each dataset into 70-15-15 train/validation/test splits. 
To further test on our method's performance in different dataset, we incorporated Winoground \cite{thrush_and_ross2022winoground}, a benchmark designed to test visuo-linguistic compositionality. 

To test on our framework's versatility across different pretrained backbones, we extracted text embeddings and image embeddings from wider set of models: for language models, Gemma2 (2B,9B) \cite{gemma_2024}, Llama3.2 (1B,3B)  \cite{llama3modelcard}, OLMo2 (7B,13B) \cite{olmo20242olmo2furious}, and for vision models, DINOv2 \cite{oquab2023dinov2}, MAE \cite{mae}, and Swav \cite{swav} for self-supervised learning (SSL) objective and ResNet50 \cite{he2016deep}, Swin \cite{swin}, and ViT \cite{dosovitskiy2020vit} for supervised (Sup) objective. (In Table \ref{tab:spreadloss}, we show results for DINOv2 and RestNet50 for vision backbones in Sugarcrepe setting. For full results (all possible pretrained backbones configurations and Winoground task), refer to Appendix Table \ref{tab:allresults}, \ref{tab:winoground}.)

For each task, we train our Joint Autoencoder Modulator (JAM) with Spread loss for 100 epochs with a batch size of 32, using data seeds 5, 42, and 55. Both autoencoders are optimized jointly using AdamW \cite{adamw} with gradient clipping (1.0) and a cosine annealing scheduler. The reconstruction loss is weighted by a linearly decaying factor $\lambda(t)$, decreasing from 1.0 to 0.1 over training epochs to gradually emphasize the alignment objective. Every five epochs, we compute image-to-text Recall@1 on the validation set, applying early stopping if no improvement is observed for five consecutive validations. 
We evaluate on two retrieval settings: (1) binary choice between the positive and its hard negative (standard in fine-grained evaluation \cite{hsieh2023sugarcrepe, yuksekgonul2023when}), and (2) 5-way choice including three additional distractors. For Winoground \cite{thrush_and_ross2022winoground}, we used their proposed text-score metrics for evaluation, defined as the percentage of instances where, for a fixed caption, the model assigns a higher similarity to the correct image than to the incorrect one. All reported results are averaged across three seed runs.

To contextualize our approach—designed to escape the “Platonic cave” of unimodal models through post-hoc alignment—we compare against CLIP \cite{clip}, a model natively trained for multimodal representation learning. For fairness, we also fine-tuned CLIP on the same train/val/test splits using the fine-tuning method proposed in \cite{yuksekgonul2023when}. To further motivate our JAM architecture with spread loss, we experimented on simple projection-based methods (linear/non-linear) with our Spread loss as well as JAM with spread loss without reconstruction component (spread w/o reconst.) as ablation experiment (Table \ref{tab:furtherexp}). 

\subsection{Does JAM enable escaping Plato's Cave? }

\begin{table*}[t!]
\centering
\vspace{-0.15in}
\resizebox{0.9\textwidth}{!}{
    \begin{tabular}{l|l|c||ccc}
    \toprule
    {Language Backbone } & {Vision Backbone } & {Alignment Method} & {Replace Task} & {Add Task} & {Swap Task} \\
    \midrule
    Gemma2 (2B) & DINOv2 (ViT-B; 86M) & Con & 65.14 & 58.34 & 57.36 \\
     & & NegCon & 86.32 & 96.85 & 74.5 \\
     & & Spread & 88.01 & 95.74 & 80.2 \\
    \midrule
    Gemma2 (2B) & ResNet50 & Con & 74.23 & 63.02 & 63.75 \\
     & & NegCon & 87.28 & 94.80 & 72.27 \\
     & & Spread & 87.60 & 94.81 & \textbf{82.17} \\
    \midrule
    Llama3.2 (1B) & DINOv2 (ViT-B) & Con & 66.97 & 60.99 & 66.59 \\
     & & NegCon & 83.43 & 94.25 & 68.26 \\
     & & Spread & 87.21 & 97.83 & 79.05 \\
    \midrule
    Llama3.2 (1B) & ResNet50 & Con & 67.36 & 67.94 & 71.11 \\
     & & NegCon & 85.74 & 94.10 & 71.69 \\
     & & Spread & 87.17 & 96.85 & \textbf{82.17} \\
    \midrule
    OLMo2 (7B) & DINOv2 (ViT-B) & Con & 68.48 & 65.17 & 66.88 \\
     & & NegCon & 85.30 & 96.67 & 71.69 \\
     & & Spread & \textbf{88.32}& 97.41 & 77.62 \\
    \midrule 
    OLMo2 (7B) & ResNet50 & Con & 68.01 & 64.35 & 73.38 \\
     & & NegCon & 89.69 & 90.12 & 71.38 \\
     & & Spread & 86.90  & \textbf{98.44} & 80.46 \\
    \midrule
    \multicolumn{3}{c||}{Pretrained CLIP (ViT-B-32/OpenAI) \cite{openclip}} & 81.05 & 77.58 & 64.69 \\
    \midrule
    \multicolumn{3}{c||}{Pretrained CLIP (ViT-B-32/LAION-400m) \cite{openclip, openclip_laion}} & 80.90 & 79.64 & 67.30 \\ 
    \midrule
    \multicolumn{3}{c||}{Pretrained CLIP (ViT-B-32/SigLIP) \cite{zhai2023sigmoid}} & 85.01 & 86.56 & 70.97 \\ 
    \midrule
    \multicolumn{3}{c||}{Finetuned CLIP (ViT-B/32)} & 74.62 & 92.69 & 67.21 \\ 
    \bottomrule
    \end{tabular}
}
    \caption{\scriptsize Image-to-Text Retrieval Results of Joint Autoencoder Modulator (JAM) for Vision-Language Compositionality.  We report Recall@1 scores (binary) across three compositional tasks in Sugarcrepe: Replace, Add, and Swap, using models trained with different alignment methods in JAM framework. JAM with Spread loss consistently outperforms contrastive baselines across all tasks and backbones. Moreover, it matches or surpasses several strong pretrained and finetuned CLIP variants, highlighting the effectiveness of structured alignment over independently pretrained representations for fine-grained vision–language reasoning.}
    \label{tab:spreadloss}
\end{table*}

\begin{table*}[t!]
\vspace{-0.1in}
\centering
\resizebox{0.75\textwidth}{!}{
    \begin{tabular}{l|l||ccc}
    \toprule
    {Post-hoc Alignment Method} & {FLOPs(G)} & {Replace Task} & {Add Task} & {Swap Task} \\
    \midrule
    Linear Proj. with Spread & 0.06 & 85.95 & 91.94 & 63.33 \\
    NonLinear Proj. with Spread & 0.12 & 86.99 & 92.22 & 68.57 \\
    JAM with Spread \textit{w/o reconst.} & 3.70 & 83.11 &  92.53 & 74.50 \\
    \textbf{JAM with Spread} & 3.70 & 88.01 & 95.74 & 80.2 \\
    \midrule
    Finetuned CLIP (ViT-B/32) & 10.48 & 74.62 & 92.69 & 67.21 \\
    \bottomrule
    \end{tabular}
}
    \caption{\scriptsize Image-to-Text Retrieval Results in Vision-Language Compositionality with Different Post-hoc Alignment architectures (Linear Projection, Nonlinear Projection, Joint Autoencoder Modulator (JAM)) with Spread Loss and ablation result for removing the reconstruction component in Spread Loss. Same evaluation scheme as shown in Table \ref{tab:spreadloss}. We show results using Gemma2 (2B) and DINOv2 (ViT-B) as language and vision backbone. We also provide FLOPs(G) across alignment methods (including finetuning CLIP) to show that JAM with Spread framework shows the best performance but requires less FLOPs compared to finetuning.}
    \label{tab:furtherexp}
    \vspace{-0.3in}
\end{table*}

Our experiment results (Table \ref{tab:spreadloss} for Sugarcrepe with binary Recall@1 setting, Table \ref{tab:allresults} in Appendix for Sugarcrepe with all pretrained backbone configurations with binary \& 5-way Recall@1 scores, Table \ref{tab:winoground} in Appendix for Winoground task) show that the proposed Spread loss consistently outperforms both the Con and NegCon baseline and pretrained/finetuned CLIP across tasks. Spread's superior performance can be attributed to its structured supervision, which encourages embeddings to reflect local clusters and semantic proximity, thereby enabling a more fine-grained alignment. This contrasts with other methods that rely solely on pairwise correspondence.

Our method achieves this competitive performance using a lightweight, post-hoc alignment approach with a joint autoencoder, which avoids the need for massive paired datasets and architectural complexity associated with pretraining large-scale multiomodal models like CLIP. This makes our method more robust in low-resource or specialized data domains, where fine-tuning a model like CLIP can lead to overfitting (i.e., losing CLIP's broad priors) and a "representation collapse" \cite{repcollapse}, as we also observe in our experiments where fine-tuning CLIP led to lower performance \ref{tab:spreadloss}. 

We also conducted stronger baseline experiments using linear and nonlinear projections over frozen embeddings to motivate our JAM framework and ablation experiment of reconstruction component in our spread loss formulation. Note that for projection-based baselines, the reconstruction component of our Spread loss is inherently absent. As Table \ref{tab:furtherexp} shows, our method (JAM+Spread) consistently outperformed projection baselines: while projections achieved lower FLOPs, they suffered from significantly lower accuracy and unstable learning due to the lack of reconstruction. Furthermore, in ablation experiments, removing the reconstruction term led to rapid overfitting, with validation loss rising sharply and models requiring early stopping to avoid collapse. This highlights the critical role of reconstruction as a regularizer, preserving generalizable structure across modalities beyond simple alignment.

\begin{figure}[!htbp]
    \centering
    \makebox[\linewidth]{\includegraphics[width=1.1\textwidth]{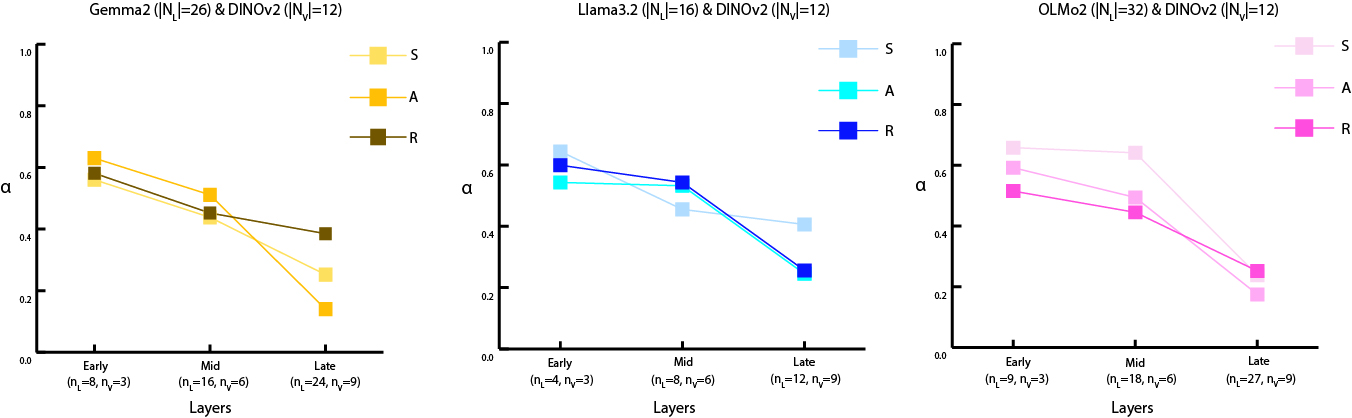}}
    \caption{\scriptsize $\alpha$ supervision with respect to the extracted embeddings layers to achieve the best retrieval accuracy for each task. $|N_L|, |N_V|$ refer to the total layers of each pretrained language, and vision model. $n_L, n_V$ refer to the layer-depth used for Early, Mid, Late experiments, respectively.}
    \label{fig:layers}
    
    \makebox[\linewidth]{\includegraphics[width=1.1\textwidth]{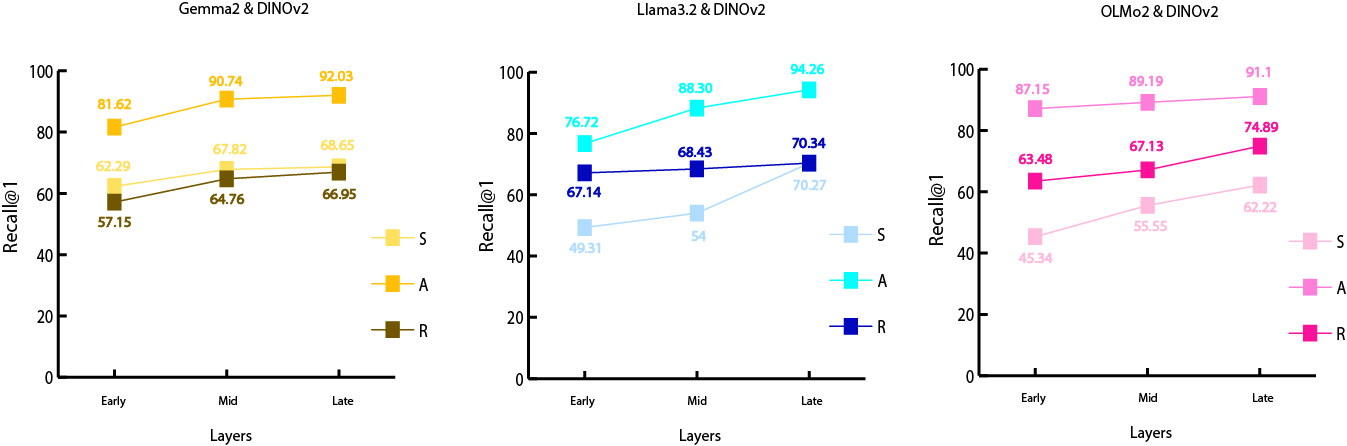}}
    \caption{\scriptsize Image-to-Text Retrieval Recall@1 (5 options case) achieved through the $\alpha$ in Figure \ref{fig:layers}. Same layer-depth configuration as Figure \ref{fig:layers}. Despite the decreasing need of context-aware supervision through  $\alpha$ in $\mathcal{L}_{spread}$, performance increases as layers progress.} \label{fig:layers_accuracy}
\end{figure}

\subsection{What kind of supervision is useful for different layer depth?: Layer-Wise Probing via Curriculum Learning} 

The previous formulation (Table \ref{tab:spreadloss}) uses a fixed $\alpha=0.5$, allowing our jointly trained text and image autoencoders to equally benefit from both components of the $\mathcal{L}_{spread}$ — the context-based alignment across modalities and the fine-grained contrastive supervision within semantically similar groups. While this balanced configuration offers strong performance across tasks, it treats the trade-off between coarse and fine supervision as static. However, the optimal degree of contextual separation may not be uniform across all settings — in particular, it may vary depending on the stage of representation (i.e., intermediate layers of pretrained backbones) used from each modality.

To investigate this, we move beyond a fixed-$\alpha$ formulation and adopt curriculum learning \cite{cl} as a probing framework (Fig. \ref{fig:layers}, \ref{fig:layers_accuracy}). Curriculum learning \cite{cl} is a training paradigm inspired by human learning, where models are first exposed to easier concepts or objectives and gradually introduced to harder or more nuanced distinctions. The motivation is to guide optimization along a smoother path: by focusing early training on easier, more generalizable patterns, the model establishes a robust representational foundation before facing fine-grained or ambiguous cases. We use curriculum learning as a diagnostic framework to probe the nature of supervision required for aligning multimodal embeddings. Specifically, we vary the balancing parameter $\alpha$ in $\mathcal{L}_{spread}$ to control the emphasis between broad semantic alignment (ConCon) and fine-grained contrast (ContextNCE). 

\vspace{-0.1in}
\paragraph{On the inverse relationship between $\alpha$ and separability}
By applying our curriculum across embeddings from early, mid, and late layers of vision and language models, we find that earlier layers consistently require higher $\alpha$ for optimal alignment—the point of peak retrieval accuracy (Fig. \ref{fig:layers}, \ref{fig:layers_accuracy}). This mirrors interpretability studies (e.g., TCAV \cite{kim2018interpretabilityfeatureattributionquantitative}, Network Dissection \cite{netdissect2017}) showing that lower layers encode primitive or entangled features, demanding stronger supervision to isolate meaningful concepts. Here, larger $\alpha$ provides the finer discriminative signal needed to disentangle and align these early representations. Later layers, already more abstract and task-aligned, align effectively with weaker supervision. Retrieval performance also improves steadily from early to late layers, indicating that deeper layers both require less contrastive pressure and yield more directly alignable representations. Thus, the curriculum offers a controlled lens into the supervision needed across representational depth, revealing how independently trained vision and language models organize information.

\subsection{Does JAM performance scale with backbone models' scale?}

\begin{table*}[t!]
\centering
\resizebox{0.75\textwidth}{!}{
    \begin{tabular}{l|l|c||ccc}
    \toprule
    {Language Backbone } & {Vision Backbone } & {Alignment Method} & {Replace Task} & {Add Task} & {Swap Task}\\
    \midrule
    Gemma2 (9B) & DINOv2 (ViT-L; 300M) & Con & 68.77 & 59.31 & 60.92 \\
     & & NegCon & 87.52 & 98.89 & 84.44 \\
     & & Spread & 89.66 & 98.89 & 84.44 \\
    \midrule
    Llama3.2 (3B) & DINOv2 (ViT-L) & Con & 65.24 & 62.38  & 73.12 \\
     & & NegCon & 84.54 & 92.61 & 75.93 \\
     & & Spread & 89.43 & 95.74 & 82.44 \\
    \midrule
    OLMo2 (13B) & DINOv2 (ViT-L) & Con & 65.86 & 59.21 & 62.28 \\
     & & NegCon & 89.36 & 93.7 & 78.89 \\
     & & Spread & 90.53 & 97.96 & 84.71 \\
    \bottomrule
    \end{tabular}
}
    \caption{\scriptsize Model Scaling Experiment. We report image-to-text Recall@1 scores across three spatial reasoning tasks (Replace, Add, Swap) using language backbones of increasing scale—Gemma2 (9B), LLaMA 3.2 (3B), and OLMo2 (13B)—within our JAM framework. Despite varying parameter counts, we observe no consistent trend of performance improvement with model size. Spread loss performs robustly across all scales, suggesting that for fine-grained, low-data tasks, representational alignment depends more on objective design than on model scale.} 
    \label{tab:scale}
    \vspace{-0.2in}
\end{table*}

We test whether larger pretrained language backbones improve alignment in JAM, using Gemma2 (9B), LLaMA 3.2 (3B), and OLMo2 (13B) with a fixed vision encoder (DINOv2 ViT-L). The aim is to assess how representational capacity affects fine-grained multimodal alignment under different contrastive objectives.  

As shown in Table~\ref{tab:scale}, scaling language models does not consistently improve performance. Spread loss remains robust, but gains from larger models are minimal. This plateau likely reflects the small training regime (200–1000 image–text pairs), which limits the gradient signal needed for large models to leverage their capacity. Moreover, the fine-grained nature of spatial reasoning tasks emphasizes precise local alignment (e.g., prepositions, swapped attributes), reducing the advantage of broad generalist knowledge from web-scale pretraining. These results highlight the effectiveness of task-specific Spread loss in capturing nuanced multimodal contrasts.

\section{Conclusion / Future Directions}
Through this framework, we provide empirical evidence and insight into the nature of representational convergence. We show that, despite originating from disjoint modalities and being trained independently, unimodal representations can be aligned through post hoc joint autoencoding — revealing a Platonic representation that supports cross-modal coherence. We propose
a practical model training recipe: the Joint Autoencoder Modulator (JAM), a Pareto-efficient framework for building specialist multimodal models on top of frozen unimodal foundations. 
Our findings show that 
post-hoc alignment via lightweight adaptation and structured supervision (i.e., $\mathcal{L}_{spread}$) can rival or even outperform generalist architectures in specialist settings. Looking forward, we see JAM as a flexible testing ground for modality alignment under varying supervision regimes. 

There are many directions for future work, including extending to domains that demand either highly specific reasoning (e.g., legal, medical, scientific), scaling to larger datasets and models, as well as exploring more sophisticated alignment approaches beyond our Spread loss.  At larger scales, it becomes interesting to also compare accuracy-efficiency tradeoffs versus other approaches such as adapter-based and joint-training approaches.  Additionally, studying these questions for more than two modalities simultaneously can be interesting, such as for time series \cite{liang2024foundation,talukdertotem,goswami2024moment,das2024decoder}.

\clearpage
\phantomsection

\appendix
\renewcommand{\thesection}{\Alph{section}}  
\renewcommand{\thesubsection}{\thesection.\arabic{subsection}}  
\renewcommand{\thesubsubsection}{\thesubsection.\arabic{subsubsection}}  
\addcontentsline{toc}{section}{Appendix Contents}
\tableofcontents

\clearpage
\section{Impact Statement} 

\textbf{Platonic Alignment: Escaping the Plato's Cave.} 

Platonic Representation Hypothesis \cite{platonic} suggests that vision and language models, though trained on disjoint objectives and data, encode latent geometries that are mutually compatible.  Our work turns this philosophical and observational claim into practice: we \emph{escape Plato’s Cave} by \emph{surfacing shared structure from representations originally confined to separate unimodal worlds}, while keeping every pretrained unimodal backbones completely frozen.

\textbf{Joint Autoencoder Modulator (JAM).} 

We propose JAM, a post‑hoc adaptor for platonic alignment that:
\vspace{-0.5em}
\begin{itemize}[leftmargin=*, itemsep=2pt]
\item \emph{Preserves specialization} – Modality‑specific autoencoders reconstruct each latent space, guarding task‑relevant features. 
\item \emph{Enforces cross‑modal coherence} – Our novel spread loss couples the JAM framework, amplifying the latent commonality predicted by the Platonic hypothesis.
\item \emph{Light-weight framework} – no end‑to‑end multimodal fine‑tuning (nor pre-training); refer to \ref{subsec:jam_params} that shows the lightweight structure of JAM.  
\end{itemize}

\textbf{Contributions.}
\vspace{-0.5em}
\begin{itemize}[leftmargin=*, itemsep=2pt]
\item \emph{Platonic Alignment through Specialist Task Settings}: We test the platonic alignment through both statistical and model-training framework in specialist task settings which require more fine‑grained contextual understanding of the data.  
\item \emph{Revisiting multimodal training pipeline strategy}: Our results show that aligning the frozen pretrained backbones can rival with inherently multimodal models, charting a potential alternative strategy for creating multimodal models for specialized task frameworks.  
\end{itemize}

\section{Statistical Metrics to Measure Platonic Alignment} 

\FloatBarrier
\subsection{Thematic analysis of statistical representation alignment tests}
We focus on the following four statistical metrics for testing Platonic alignment: Canonical Correlation Analysis (CCA \cite{cca}) with linear and kernel (rbf kernel) PCA, Singular Value CCA (SVCCA \cite{svcca}), Centered Kernel Alignment (CKA \cite{cka}), and Centered Kernel Nearest Neighbors Alignment (CKNNA \cite{nearest_neighbors, platonic}). 

The following are the thematic analyses that demonstrate our metrics selection strategy: 
\vspace{-0.5em}
\begin{itemize}[leftmargin=*, itemsep=4pt]
\item \textbf{Unsupervised and second-order:} All four metrics are \emph{unsupervised} and based on second-order statistics (e.g., inner products or covariances), making them scale-invariant and naturally comparable across modalities.
\item \textbf{Symmetricity:} The symmetry property indicates that the metric treats the data points interchangeably, meaning $d(x,y)=d(y,x))$.
\item \textbf{Data-Driven (D) vs.\ Canonical (C):}
\begin{itemize}[leftmargin=1.5em]
    \item \textbf{Data-Driven} metrics (CCA, SVCCA) learn linear projections from the data that actively mold one representation space to align with another. They measure the \emph{best-case alignment} achievable through adaptation.
    \item \textbf{Canonical} metrics (CKA, CKNNA) keep the input representations fixed and evaluate the \emph{existing alignment structure} as is, without transformation. They quantify inherent similarity without retraining or reprojecting.
\end{itemize}
\item \textbf{Global (G) vs.\ Local (L):}
\begin{itemize}[leftmargin=1.5em]
    \item \textbf{Global} metrics (CCA, SVCCA, CKA) aggregate information across \emph{all sample pairs}, capturing holistic alignment patterns between entire distributions.
    \item \textbf{Local} metrics (CKNNA) focus on \emph{localized structure} by comparing only mutual $k$-nearest neighbors, revealing fine-grained neighborhood-level similarity that may be obscured globally.
\end{itemize}
\item \textbf{Batchable:} A computational property that makes it feasible to compute in a reasonable time frame.
\end{itemize}

\begin{table*}[t]
\centering
\caption{Comparative analysis of statistical alignment metrics. All metrics are unsupervised and based on second-order statistics.}
\label{tab:metric_property_matrix}
\renewcommand{\arraystretch}{1.2}
\setlength{\tabcolsep}{5pt}
\scriptsize
\begin{tabular}{l cccc m{3.5cm}}
\toprule
\textbf{Metric} & \multicolumn{4}{c}{\textbf{Property}} & \textbf{Description} \\
\cmidrule(lr){2-5}
& \textbf{Symmetric} & \textbf{Data-Driven(D) vs. Canonical(C)} & \textbf{Global(G) vs. Local(L)} & \textbf{Batchable} & \\
\midrule
CCA & \cmark & D & G & \cmark & Learns projections to maximize cross-view correlation. Sensitive to high-dimensional data structure, hence apply linear or kernel PCA before applying CCA. \\
\lightmidrule
SVCCA & \cmark & D & G & \cmark & Applies SVD to smooth and compress representations before applying CCA. \\
\lightmidrule
CKA & \cmark & C & G & \cmark & Compares original representations using kernel alignment, invariant to rotation and scale. \\
\lightmidrule
CKNNA & \cmark & C & L & \cmark & A local variant of CKA restricted to shared $k$-nearest neighbors; reveals localized alignment structure. \\
\bottomrule
\end{tabular}
\end{table*}

\FloatBarrier
\subsection{Metrics Formulation}
Recall our glossary: 
\begin{table}[h!]
\vspace{-0.15in}
\scriptsize
\caption{Glossary of variables and symbols}
\label{glossary}
\centering
    \begin{tabularx}{15cm}{cl}
    \hline
    Symbol & Description \\ 
    \hline
    $\Phi_V, \Psi_V$ & Vision (image) autoencoder (encoder, decoder) \\
    $\Phi_L, \Psi_L$ & Language (text) autoencoder (encoder, decoder) \\
    $V$ & Set of image embeddings ($|V| = n, v_i \in V, i=\{1,\dots, n\}$) \\
    $L$ & Set of text embedding (i.e., $L_P \cup L_N$; $|L| = 2n$) \\
    $L_P$ & Set of positive text embeddings ($|L_P| = n, l_{P_i} \in L_P$) \\
    $L_N$ & Set of \textbf{hard} negative text embeddings ($|L_N| = n, l_{N_i} \in L_N$) \\ 
    $d_{input}, d_{latent}$ & Input, Latent (i.e., bottleneck) dimension of autoencoder \\ 
    $l_i$ & Element of $L \setminus \{l_{P_i}\}$ with anchor index $i$ \\ 
    $C_L(i)$ & Set of similar context text embeddings \\
    $\widetilde{C_L}(i)$ & Set of dissimilar context text embeddings (i.e., $L \setminus C_L(i))$ \\
    \hline
    \end{tabularx}
\end{table}  

\paragraph{Kernel preliminaries}
Let $\{(\mathbf x_i,\mathbf y_i)\}_{i=1}^{n}$ be paired samples: in our case, let $x_i$ be the images and $y_i$ be the corresponding texts. A representation function $f_V:\mathcal X\!\to\!V\in\mathbb R^{d}$ maps inputs to feature vectors $v_i=f_V(\mathbf x_i)$ in an RKHS (Reproducing Kernel Hilbert Space) $\mathcal H$ \cite{rkhs} equipped with the inner‑product kernel $K_{ij}=\kappa(v_i,v_j)=\langle v_i,v_j \rangle$. Analogously $f_L$ maps $\mathbf y_i$ to $l_i$ with kernel $L_{ij}=\langle l_i,l_j \rangle$. Center both kernels to remove mean effects via the centring matrix $H=I-\frac1n\mathbf 1\mathbf 1^{\!\top}$: \[\bar K = HKH, \qquad \bar L = HLH.\]

\subsubsection{Centered‑Kernel Alignment (CKA)}
CKA \cite{cka} normalises the \emph{Hilbert–Schmidt Independence Criterion} (HSIC) \cite{hsic} between
the centred kernels:%
\begin{align}
\operatorname{HSIC}(K,L)
      &=\frac1{(n-1)^2}\operatorname{tr}\bigl(\bar K\,\bar L\bigr), 
      \label{eq:hsic} \\
\operatorname{CKA}(K,L)
      &=\frac{\operatorname{HSIC}(K,L)}{\sqrt{\operatorname{HSIC}(K,K)\, \operatorname{HSIC}(L,L)}}.
      \label{eq:cka}
\end{align}

Expanding the trace in~\eqref{eq:hsic} shows that every pair $(i,j)$ contributes a product of centered similarities:
\begin{align}
\operatorname{tr}(\bar K\bar L)=
\sum_{i=1}^{n}\sum_{j=1}^{n}
\bigl(\langle v_i,v_j \rangle-\mathbb E_\ell\langle v_i,v_\ell \rangle\bigr)
\bigl(\langle l_i,l_j \rangle-\mathbb E_\ell\langle l_i,l_\ell \rangle\bigr). 
\end{align}
Because the denominator applies the same operation to each modality, CKA is scale‑invariant and bounded in $[0,1]$; it therefore has advantage in measuring \emph{global} geometric correspondence. However, equation~\eqref{eq:cka} aggregates \emph{all} pairwise relations, so even a small region of mis‑aligned samples suppresses the score which CKA can be too strict for measuring cross-modal cases. 

\subsubsection{Centered‑Kernel \texorpdfstring{$k$}{k}-NN Alignment (CKNNA)}
To focus on \emph{local} structure, the authors in \cite{nearest_neighbors, platonic} suggest a binary mask~$\alpha(i,j)$ that selects only pairs that are \emph{mutual} $k$‑nearest neighbours in \emph{\textbf{both} modalities}: 
\vspace{0.5em}
\begin{align}
\alpha(i,j) &=
\mathbbm{1} \!\bigl[
      v_j\!\in\!\text{kNN}(v_i)\ \wedge\ 
      l_j\!\in\!\text{kNN}(l_i)\ \wedge\ i\neq j
\bigr].
\end{align}

Using this mask, one can restrict the definition of alignment to bias more towards the local structure through replacing the full HSIC trace by a \emph{masked} cross‑covariance.:
\begin{align}
\operatorname{Align}_{\text{local}}(K, L)
    &= \sum_{i=1}^{n}\sum_{j=1}^{n} \alpha(i,j)
    \bigl(\langle v_i,v_j \rangle-\mathbb E_\ell\langle v_i,v_\ell \rangle\bigr)
    \bigl(\langle l_i,l_j \rangle-\mathbb E_\ell\langle l_i,l_\ell \rangle\bigr) \\
    &=\sum_{i=1}^{n}\sum_{j=1}^{n}\alpha(i,j)\,\bar K_{ij}\,\bar L_{ij} \label{eq:tr_knn}. 
\end{align}
The final formulation of CKNNA becomes: 
\vspace{0.5em}
\begin{align}
\operatorname{CKNNA}(K,L)
      &=\frac{\operatorname{Align}_{\text{local}}(K,L)}
             {\sqrt{\operatorname{Align}_{\text{local}}(K,K)\, \operatorname{Align}_{\text{local}}(L,L)}}. \label{eq:cknna}
\end{align}

Note that $\operatorname{CKNNA}$ can be fully recovered to $\operatorname{CKA}$ for $k\!\to\! n$, since $\alpha(i,j)\!=\!1$ for all $i\!\neq\!j$. 

Thus, CKA asks: \emph{“Are the two representations \textbf{globally} linearly related?”}; whereas CKNNA captures: \emph{“Do the two models agree \textbf{locally} (i.e., on each point’s neighbours)?”}.

\subsubsection{CCA (Canonical Correlation Analysis)} 
Let $X\!\in\!\mathbb R^{d_x\times n}$ and $Y\!\in\!\mathbb R^{d_y\times n}$ be column‑wise zero-centered feature matrices produced by two representations (all means are removed so that covariance estimates are unbiased).  
Define sample covariance blocks: 
\begin{align}
\Sigma_{xx}= \tfrac1nXX^{\!\top},\quad
\Sigma_{yy}= \tfrac1nYY^{\!\top},\quad
\Sigma_{xy}= \tfrac1nXY^{\!\top}.
\end{align}

CCA searches for weight vectors $\mathbf w\!\in\!\mathbb R^{d_x}$, $\mathbf v\!\in\!\mathbb R^{d_y}$ that maximise the \emph{Pearson correlation}
between projected variables:
\begin{align}
\!\!\rho
  &=\max_{\mathbf w,\mathbf v}
      \;\; \frac{\mathbf w^{\!\top}\Sigma_{xy}\mathbf v}
                  {\sqrt{\mathbf w^{\!\top}\Sigma_{xx}\mathbf w}
                         \sqrt{\mathbf v^{\!\top}\Sigma_{yy}\mathbf v}}
      \\
  &\text{s.t.}\quad
       \mathbf w^{\!\top}\Sigma_{xx}\mathbf w = 1,\;
       \mathbf v^{\!\top}\Sigma_{yy}\mathbf v = 1.
\end{align}

Using Lagrange multipliers we obtain the generalized eigenvalue problem: 
\begin{align}
\bigl(\Sigma_{xx}^{-1}\Sigma_{xy}\Sigma_{yy}^{-1}\Sigma_{yx}\bigr)
          \mathbf w
     \;=\;\rho^{2}\,\mathbf w,
\quad
\mathbf v=\rho^{-1}\Sigma_{yy}^{-1}\Sigma_{yx}\mathbf w,
\end{align}
whose top $k$ eigenvalues $\rho_1\!\ge\!\rho_2\!\ge\!\dots\!\ge\!\rho_k$ are the \textbf{canonical correlations}. We report the maximum canonical correlation value (obtained from the first canonical component) as an alignment score. 
\paragraph{PCA‑assisted CCA}
Since CCA method suffers from high-dimensionality data structures \cite{highdcca}, we first project each feature vectors onto its leading $r$ principal components:
\begin{align}
\tilde X = P_x^{\!\top} X,\qquad
\tilde Y = P_y^{\!\top} Y,
\end{align}
with $P_x\!\in\!\mathbb R^{d_x\times r}$ (orthonormal) obtained from linear PCA or kernel PCA using an RBF kernel
$\kappa(\mathbf z,\mathbf z')=\exp[-\gamma\|\mathbf z-\mathbf z'\|^{2}]$. CCA is then run on $\tilde X,\tilde Y$, producing more stable
correlations in high‑dimensional regimes. In our experiments, we used $r=50$ (i.e., feature dimension = 50), and $k=50$ (i.e., number of canonical components = 50). 

\subsubsection{SVCCA (Singular Vector Canonical Correlation Analysis)} 
SVCCA \cite{svcca} replaces the heuristic PCA cut‑off with a \emph{data‑dependent approach} via SVD:
\begin{enumerate}
    \item Compute SVDs: 
          $X = U_x\Sigma_xV_x^{\!\top}$,
          $Y = U_y\Sigma_yV_y^{\!\top}$.
    \item Retain the top $r_x,r_y$ singular vectors explaining at least a fixed proportion $\eta$ (e.g.,\ $99\%$) of variance:
          \[
             \hat X = \Sigma_{x,[1:r_x]}V_{x,[1:r_x]}^{\!\top},\qquad
             \hat Y = \Sigma_{y,[1:r_y]}V_{y,[1:r_y]}^{\!\top}.
          \]
    \item Run linear CCA on $(\hat X,\hat Y)$ to obtain canonical correlations $\{\rho_i\}_{i=1}^{k}$ with $k=\min(r_x,r_y)$.
\end{enumerate}
The SVCCA score is reported using mean of these $k$ correlations, \(\text{SVCCA}=\tfrac1k\sum_{i=1}^{k}\rho_i,\) where we report SVCCA score with $k=10$.

\FloatBarrier
\subsection{Supplemental Results for Statistical Alignment Tests} 
We describe the specific data pair set-up and provide results of statistical alignment tests for each specific task (i.e., Replace, Add, and Swap cases) in Sugarcrepe data \cite{hsieh2023sugarcrepe}. In our experiments, we reduced features dimension to 50 for CCA using linear and kernel PCA; for SVCCA, CKA, and CKNNA, the features dimension was set to 10, and nearest neighbors $k=10$ (same setting utilized in \cite{platonic}). 

\subsubsection{Replace Task}
\begin{itemize}[leftmargin=*, itemsep=0pt]
    \item Match: $V$ and $L$ that are matching (i.e., correct correspondence) 
    \item Easy Non-Match: $V$ and $L$ that have clear non-matching aspects 
    \begin{itemize}
    {\footnotesize
        \item Case 1: $V$ = White-noise image embeddings, $L$ = Same $L$ from Match case
        \item Case 2: $V$ = Same $V$ from Match case, $L$ = Text embeddings extracted from "The Great Gatsby" novel
    }
    \end{itemize}
    \item Hard Non-Match: $V$ and $L$ that are non-matching due to specific \textit{attribute, object, and relation} being \textit{replaced} by incorrect text; hence the overarching context of this text is similar to correct matching text, but only differ by the specific aspect. Thus, more fine-grained understanding is required to discern that it is actually a non-matching text for the image. 
\end{itemize}


\begin{table*}[h] 
\centering
\caption{Replace Task: Statistical Alignment Test Results}
\label{tab:replace_stats}
\vspace{0.5em}
\resizebox{\textwidth}{!}
{\large
\setlength{\tabcolsep}{6pt}
\renewcommand{\arraystretch}{1.2}  
\begin{tabular}{l||cccc|cccc|cccc}
\toprule
& \multicolumn{4}{c|}{\cellcolor[HTML]{abebc6}\textbf{Match}} 
& \multicolumn{4}{c|}{\cellcolor[HTML]{f0b27a}\textbf{Easy Non-Match}} 
& \multicolumn{4}{c}{\cellcolor[HTML]{ec7063}\textbf{Hard Non-Match}} \\
\textbf{Model Pairs} & \textbf{CCA (linear/kernel pca)} & \textbf{CKA} & \textbf{SVCCA} & \textbf{CKNNA} 
& \textbf{CCA} & \textbf{CKA} & \textbf{SVCCA} & \textbf{CKNNA} 
& \textbf{CCA} & \textbf{CKA} & \textbf{SVCCA} & \textbf{CKNNA} \\
\midrule
Gemma2 \& DINOv2 & 0.94 / 0.94 & 0.286 & 0.432 & 0.209 & 0.48 / 0.47 & 0.020 & 0.100 & 0.006 & 0.94 / 0.94 & 0.293 & 0.403 & 0.194 \\
Llama3.2 \& DINOv2 & 0.94 / 0.95 & 0.447 & 0.425 & 0.275 & 0.47 / 0.47 & 0.027 & 0.097 & 0.010 & 0.94 / 0.94 & 0.431 & 0.444 & 0.240 \\
OLMo2 \& DINOv2 & 0.95 / 0.95 & 0.439 & 0.505 & 0.287 & 0.47 / 0.47 & 0.028 & 0.102 & 0.008 & 0.95 / 0.95 & 0.422 & 0.482 & 0.252 \\
Gemma2 \& ResNet50 & 0.91 / 0.91 & 0.25 & 0.430 & 0.190 & 0.51 / 0.51 & 0.028 & 0.100 & 0.006 & 0.91 / 0.91 & 0.250 & 0.420 & 0.190 \\ 

\bottomrule
\end{tabular}
}
\end{table*}

\subsubsection{Add task}
\begin{itemize}[leftmargin=*, itemsep=0pt]
    \item Match: $V$ and $L$ that are matching (i.e., correct correspondence) 
    \item Easy Non-Match: $V$ and $L$ that have clear non-matching aspects 
    \begin{itemize}
    {\footnotesize
        \item Case 1: $V$ = White-noise image embeddings, $L$ = Same $L$ from Match case
        \item Case 2: $V$ = Same $V$ from Match case, $L$ = Text embeddings extracted from "The Great Gatsby" novel
    }
    \end{itemize}
    \item Hard Non-Match: $V$ and $L$ that are non-matching due to specific \textit{attribute, object} being \textit{added}, which leads to incorrect correspondence of reality
\end{itemize}


\begin{table*}[h] 
\centering
\caption{Add Task: Statistical Alignment Test Results}
\label{tab:add_stats}
\vspace{0.5em}
\resizebox{\textwidth}{!}
{\large
\setlength{\tabcolsep}{6pt}
\renewcommand{\arraystretch}{1.2} 
\begin{tabular}{l||cccc|cccc|cccc}
\toprule
& \multicolumn{4}{c|}{\cellcolor[HTML]{abebc6}\textbf{Match}} 
& \multicolumn{4}{c|}{\cellcolor[HTML]{f0b27a}\textbf{Easy Non-Match}} 
& \multicolumn{4}{c}{\cellcolor[HTML]{ec7063}\textbf{Hard Non-Match}} \\
\textbf{Model Pairs} & \textbf{CCA} & \textbf{CKA} & \textbf{SVCCA} & \textbf{CKNNA} 
& \textbf{CCA} & \textbf{CKA} & \textbf{SVCCA} & \textbf{CKNNA} 
& \textbf{CCA} & \textbf{CKA} & \textbf{SVCCA} & \textbf{CKNNA} \\
\midrule
Gemma2 \& DINOv2 & 0.95 / 0.94 & 0.311 & 0.425 & 0.201 & 0.48 / 0.47 & 0.022 & 0.095 & 0.007 & 0.94 / 0.94 & 0.323 & 0.448 & 0.206 \\
Llama3.2 \& DINOv2 & 0.95 / 0.95 & 0.451 & 0.424 & 0.257 & 0.48 / 0.48 & 0.027 & 0.097 & 0.011 & 0.95 / 0.94 & 0.444 & 0.447 & 0.259 \\ 
OLMo2 \& DINOv2 & 0.94 / 0.94 & 0.448 & 0.491 & 0.269 & 0.47 / 0.47 & 0.028 & 0.097 & 0.009 & 0.94 / 0.94 & 0.448 & 0.481 & 0.265 \\
Gemma2 \& ResNet50 & 0.89 / 0.89 & 0.309 & 0.420 & 0.199 & 0.51 / 0.51 & 0.028 & 0.100 & 0.006 & 0.89 / 0.89 & 0.320 & 0.400 & 0.200\\
\bottomrule
\end{tabular}
}
\end{table*}

\subsubsection{Swap task}
\begin{itemize}[leftmargin=*, itemsep=0pt]
    \item Match: $V$ and $L$ that are matching (i.e., correct correspondence) 
    \item Easy Non-Match: $V$ and $L$ that have clear non-matching aspects 
    \begin{itemize}
    {\footnotesize
        \item Case 1: $V$ = White-noise image embeddings, $L$ = Same $L$ from Match case
        \item Case 2: $V$ = Same $V$ from Match case, $L$ = Text embeddings extracted from "The Great Gatsby" novel
    }
    \end{itemize}
    \item Hard Non-Match: $V$ and $L$ that are non-matching due to specific \textit{attribute, object} being \textit{swapped}
\end{itemize}


\begin{table*}[h] 
\centering
\caption{Swap Task: Statistical Alignment Test Results}
\label{tab:swap_stats}
\vspace{0.5em}
\resizebox{\textwidth}{!}
{\large
\setlength{\tabcolsep}{6pt} 
\renewcommand{\arraystretch}{1.2}  
\begin{tabular}{l||cccc|cccc|cccc}
\toprule
& \multicolumn{4}{c|}{\cellcolor[HTML]{abebc6}\textbf{Match}} 
& \multicolumn{4}{c|}{\cellcolor[HTML]{f0b27a}\textbf{Easy Non-Match}} 
& \multicolumn{4}{c}{\cellcolor[HTML]{ec7063}\textbf{Hard Non-Match}} \\
\textbf{Model Pairs} & \textbf{CCA} & \textbf{CKA} & \textbf{SVCCA} & \textbf{CKNNA} 
& \textbf{CCA} & \textbf{CKA} & \textbf{SVCCA} & \textbf{CKNNA} 
& \textbf{CCA} & \textbf{CKA} & \textbf{SVCCA} & \textbf{CKNNA} \\
\midrule
Gemma2 \& DINOv2 & 0.95 / 0.95 & 0.317 & 0.440 & 0.228 & 0.47 / 0.47 & 0.037 & 0.124 & 0.014 & 0.94 / 0.94 & 0.330 & 0.433 & 0.220 \\
Llama3.2 \& DINOv2 & 0.94 / 0.94 & 0.463 & 0.473 & 0.311 & 0.47 / 0.47 & 0.049 & 0.129 & 0.012 & 0.94 / 0.94 & 0.453 & 0.430 & 0.287 \\ 
OLMo2 \& DINOv2 & 0.95 / 0.94 & 0.466 & 0.462 & 0.312 & 0.480 / 0.48 & 0.05 & 0.130 & 0.012 & 0.94 / 0.94 & 0.455 & 0.511 & 0.276 \\
Gemma2 \& ResNet50 & 0.89 / 0.89 & 0.315 & 0.430 & 0.221 & 0.50 / 0.50 & 0.029 & 0.123 & 0.014 & 0.89 / 0.89 & 0.315 & 0.428 & 0.220 \\
\bottomrule
\end{tabular}
}
\end{table*}

\FloatBarrier
\section{Spread Loss Formulation for Multimodal Framework}
\subsection{Our Formulation of Spread Loss for Multimodal Representation Learning}
\begin{equation}
    \mathcal{L}_{spread} = \frac{1}{2} \left [\mathcal{L}_{spread-\text{VL}} + \mathcal{L}_{\text{LV}} \right]
\end{equation}
The formulation for VL direction is a weighted sum with $\alpha$ controlling the trade-off between context-level alignment and intra-context contrast: 
\begin{equation}
    \mathcal{L}_{spread-\text{VL}} = (1-\alpha)\mathcal{L}_{ConCon} + \alpha\mathcal{L}_{contextNCE}
\end{equation}

For $\alpha \in [0, 1]$, per-sample case of $\mathcal{L}_{spread-\text{VL}}$ is defined as: 
\begin{equation}
    \mathcal{L}_{spread-\text{VL}}(\Phi_V, v_i, B) = (1-\alpha)\mathcal{L}_{ConCon}(\Phi_V, v_i, B) + \alpha\mathcal{L}_{contextNCE}(\Phi_V, v_i, B)
\end{equation}
where 
\begin{equation}
    \mathcal{L}_{ConCon}(\Phi_V, v_i, B) = - \frac{1}{|C_L(i)|}\sum_{c_l \in  C_L(i)} \log \left( \frac{\sigma (v_i, c_l)}{\sigma(v_i, c_l) + \sum_{\widetilde{c_l} \in \widetilde{C_L}(i)} \sigma(v_i, \widetilde{c_l})} \right) , 
    \label{eq:lconcon}
\end{equation}
\begin{equation}
    \mathcal{L}_{contextNCE}(\Phi_V, v_i, B) = - \log \left( \frac{\sigma(v_i, l_{P_i})}{\sum_{c_l \in C_L(i)} \sigma(v_i, c_l)} \right). 
    \label{eq:lcontextnce}
\end{equation}

The overall loss $\mathcal{L}_{spread-\text{VL}}(\Phi_V, B)$ is computed by averaging over all $(v_i, l_{P_i}, l_{N_i}) \in B$:
\[\mathcal{L}_{spread-\text{VL}}(\Phi_V, B) = \frac{1}{|B|} \sum_{i=1}^{|B|} \mathcal{L}_{spread-\text{VL}}(\Phi_V, v_i, B).\]

Recall that since hard-negative texts do not have correct corresponding images, the formulation for LV direction omits hard-negative texts. So, it follows the standard LV loss in contrastive learning scheme of CLIP models:
\begin{equation}
    \mathcal{L}_{\text{LV}}(\Phi_L, B) = - \frac{1}{|B|} \sum_{i=1}^{|B|} \log \frac{\sigma(l_{P_i}, v_i)}{\sum_{j=1, j \neq i}^N \sigma(l_{P_i}, v_j)}
\end{equation}

\subsection{Spread Loss Formulation in Visual Representation Learning Domain} 
Spread loss in vision representation learning domain \cite{chen2022perfectly} is constructed by a weighted sum of a supervised contrastive loss ($\mathcal{L}_{\text{sup}}$) \cite{supcon} and a class-conditional InfoNCE loss ($\mathcal{L}_{\text{cNCE}}$) \cite{clip}. The core motivation is similar in our multimodal scheme: training encoder to produce representations of the data by pulling together similar points (positive pairs) and pushing apart dissimilar points (negative pairs). 

Formally, let $B$ be a batch of data from dataset $\mathcal{D}$. Define the positive set:
\[P(i, B) = \{x^+ \in B \setminus \{x_i\} : h(x^+) = h(x_i)\}\]
and the negative set:
\[N(i, B) = \{x^- \in B \setminus \{x_i\} : h(x^-) \neq h(x_i)\},\]
where $h(x)$ denotes the class label of $x$ and $a(x_i)$ be an augmentation of $x_i$. Define the similarity (i.e., cosine similarity) with temperature hyperparameter $\tau > 0$:
\[\sigma_f(x, x') = \exp\left( \frac{f(x)^\top f(x')}{\tau} \right).\]
For $\alpha \in [0, 1]$, the per-sample spread loss is defined as:
\begin{equation}
    \mathcal{L}_{spread}(f, x_i, B) = (1 - \alpha) \mathcal{L}_{sup}(f, x_i, B) + \alpha \mathcal{L}_{cNCE}(f, x_i, B), 
\end{equation}
where
\begin{equation}
    \mathcal{L}_{sup}(f, x_i, B) = - \frac{1}{|P(i, B)|} \sum_{x^+ \in P(i, B)} \log \left( \frac{ \sigma_f(x_i, x^+) }{ \sigma_f(x_i, x^+) + \sum_{x^- \in N(i, B)} \sigma_f(x_i, x^-) } \right),
    \label{eq:lsup}
\end{equation}
\begin{equation}
    \mathcal{L}_{cNCE}(f, x_i, B) = - \log \left( \frac{ \sigma_f(x_i, a(x_i)) }{ \sum_{x^+ \in P(i, B)} \sigma_f(x_i, x^+) } \right).
    \label{eq:lcNCE}
\end{equation}

The overall loss $\mathcal{L}_{spread}(f, B)$ is computed by averaging over all $x_i \in B$:
\[\mathcal{L}_{spread}(f, B) = \frac{1}{|B|} \sum_{x_i \in B} \mathcal{L}_{spread}(f, x_i, B).\]
$\mathcal{L}_{\text{sup}}$ encourages intra-class clustering by pulling together same-class samples, while $\mathcal{L}_{\text{cNCE}}$ repels within-class samples except for augmentations. Their combination spreads same-class points while preserving instance-level attraction, promoting more structured representation spaces.

\FloatBarrier
\subsection{Side‑by‑Side Comparison}

\paragraph{Intuition.}
$\mathcal{L}_{ConCon}$ \emph{pulls} an image toward all texts that share its visual context (analogous to class‑level clustering, performed by $\mathcal{L}_{sup}$), while $\mathcal{L}_{contextNCE}$ \emph{pushes} it away from other texts inside that context except its primary caption
(analogous to instance discrimination, performed by $\mathcal{L}_{cNCE}$).

Table \ref{tab:loss_notation_compare} summarizes the symbols used in the Spread loss in vision domain (left column) and their direct counterparts in our our multimodal formulation (right column). The first row clarifies that the visual setting relies on a single encoder $f(\cdot)$, whereas the multimodal variant distinguishes between an image encoder $\Phi_V$ and a text encoder $\Phi_L$. Subsequent rows pair up the anchor, positive, and negative sets, making it explicit that class labels in the visual domain translate to visual contexts (collections of captions). Finally, the similarity function retains the same softmax‐temperature structure; only the argument types differ (image-image vs. image–text).

Table \ref{tab:loss_objective_compare} presents the four loss terms from an objective-driven perspective. Each row identifies the loss and breaks down its optimization goal into four parts: the anchor being updated, the positive samples it should align with, the negatives it should separate from, and the broader motivation behind this push-pull dynamic. This format highlights a consistent analogy: $\mathcal{L}_{sup}$ clusters images by class, while our $\mathcal{L}_{ConCon}$ clusters images with all captions from the same visual context. In contrast, $\mathcal{L}_{cNCE}$ and $\mathcal{L}_{contextNCE}$ act as sharpening losses, refining clusters by contrasting a specific positive — an augmentation for vision case and a primary caption for our multimodal case — against hard negatives. 

\begin{table}[t]
\centering
\caption{Notation Comparison of Spread Loss in Vision vs. Multimodal Domain (Ours)}
\begin{tabular}{l|c|c}
\toprule
\textbf{Symbol} & \textbf{Vision Spread} & \textbf{Multimodal Spread (Ours)}\\
\midrule
Encoders & $f(\cdot)$ & $\Phi_V(\cdot),\,\Phi_L(\cdot)$ \\
\lightmidrule
Anchor & $x_i$ (i.e., image) & $v_i=\Phi_V(x_i)$ \\
\lightmidrule
Positives & $P(i)$ same class & $C_L(i) = \{l_{P_i}, l_{N_i}\}$ (i.e., similar context texts) \\
\lightmidrule
Negatives & $N(i)$ diff. class & $\widetilde{C}_L(i) = L \setminus C_L(i)$ \\
\lightmidrule
Similarity & $\sigma(x,x')$ & $\sigma(v,l)$ \\
\bottomrule
\end{tabular}
\label{tab:loss_notation_compare}
\end{table}

\begin{table}[t]
\centering
\caption{"Objective" view of each loss term.}
\begin{tabular}{lllll}
\toprule
\textbf{Loss} & \textbf{Anchor} & \textbf{Positives} & \textbf{Negatives} & \textbf{Goal}\\
\midrule
$\mathcal{L}_{sup}$ & $x_i$  & same class & other classes & intra‑class cohesion \\
\lightmidrule
$\mathcal{L}_{ConCon}$ & $v_i$  & $C_L(i)$   & $\widetilde{C}_L(i)$ & cross‑modal cohesion \\
\midrule
$\mathcal{L}_{cNCE}$ & $x_i$  & $a(x_i)$   & other $P(i)$  & instance sharpening \\
\lightmidrule
$\mathcal{L}_{contextNCE}$ & $v_i$  & $l_{P_i}$  & $C_L(i)\!\setminus\!\{l_{P_i}\}$ & context sharpening \\
\bottomrule
\end{tabular}
\label{tab:loss_objective_compare}
\end{table}

\FloatBarrier

\section{Model Analysis \& Training Configuration}

\subsection{Pretrained Unimodal Backbones}
\begin{table}[H]
\footnotesize
\centering
\label{tab:unimodal_models}
\begin{minipage}[t]{0.43\textwidth}
\centering
\textbf{LM} \\[0.5ex] 
\begin{tabular}{llcc}
    \toprule
    \textbf{Model} & \textbf{Size} & \textbf{\# Layers} & \textbf{Dim} \\
    \midrule
    Gemma2 & 2B & 26 & 2304 \\
    & 9B & 42 & 3584 \\
    \midrule
    Llama3.2 & 1B & 16 & 2048 \\
    & 3B & 32 & 4096 \\
    \midrule
    OLMo2 & 7B & 32 & 4096 \\
    & 13B & 40 & 5120 \\
    \bottomrule
\end{tabular}
\end{minipage}
\hfill
\begin{minipage}[t]{0.51\textwidth}
\centering
\textbf{VM} \\[0.5ex] 
\begin{tabular}{llcc}
\toprule
\textbf{Model} & \textbf{Size} & \textbf{\# Layers} & \textbf{Dim} \\
\midrule
DINOv2 (SSL) & 86M (ViT-B) & 12 & 768 \\
       & 300M (ViT-L) & 12 & 1024 \\
\midrule
MAE (SSL) & 86M & 12 & 768 \\ 
\midrule
Swav (SSL) & 23M & 50 & 2048 \\
\midrule
Swin (Sup) & 88M & 50 & 2048 \\
\midrule
ViT (Sup) & 86M & 12 & 768 \\
\midrule
ResNet50 (Sup) & 23M & 50 & 2048 \\ 
\bottomrule
\end{tabular}
\end{minipage}
\vspace{+1em}
\caption{Configuration of pretrained unimodal models used in experiments. For MAE, we used ViT as backbone architecture. For Swav and Swin, we used RestNet50 as backbone architecture.}
\end{table}

\subsection{JAM (Joint Autoencoder Modulator) Framework} \label{subsec:jam_params}
We experiment JAM (Joint Autoencoder Modulator) framework with 3 hidden layers with dimension size 512, bottleneck/latent layer with dimension size 256, drop out ratio of 0.1, batch size of 32, and SwiGLU activation. With this architecture, Table \ref{tab:jam_scale_analysis} shows the model analysis of JAM attached to each pretrained unimodal backbones. 

With the extracted unimodal features (language and vision, respectively) data, for each task setting, we use 70-15-15 train/validation/test splits. For each task, we train our Joint Autoencoder Modulator (JAM) with all the loss schemes ($\mathcal{L}_{spread}$, $\mathcal{L}_{con}$, $\mathcal{L}_{NegCon}$) for 100 epochs with a batch size of 32, using data seeds 5, 42, and 55. The reported scores are the average of recall scores across different seeds. Both autoencoders are optimized jointly using AdamW \cite{adamw} with gradient clipping (1.0) and a cosine annealing scheduler. We initialize the logit scaling parameter in log-space as $log(1/0.07)$, following the common CLIP \cite{clip} initialization strategy. During training, the effective scale is recovered via exponentiation, allowing the model to start with sharper similarity distributions and learn an appropriate temperature dynamically. The reconstruction loss is weighted by a linearly decaying factor $\lambda(t)$, decreasing from 1.0 to 0.1 over training epochs to gradually emphasize the alignment objective. Every five epochs, we compute image-to-text Recall@1 on the validation set, applying early stopping if no improvement is observed for five consecutive validations. We evaluate on two retrieval settings: (1) binary choice between the positive and its hard negative (standard evaluation scheme in fine-grained task setting \cite{hsieh2023sugarcrepe, yuksekgonul2023when}), and (2) a 5-way choice including three additional distractors. 

\begin{table}[H] \label{tab:jam_scale_analysis}
\centering
\begin{tabular}{ccc}
    \toprule
    \textbf{Pretrained Backbone} & \textbf{JAM Parameters (M)} & \textbf{JAM FLOPs (G)} \\
    \midrule
    Gemma2 (2B)  & 11.55 & 2.39 \\
    Gemma2 (9B)  & 16.20 & 3.31 \\
    \midrule
    Llama3.2 (1B)  & 10.55 & 2.06 \\
    Llama3.2 (3B)  & 18.31 & 3.81 \\
    \midrule
    OLMo2 (7B)  & 18.31 & 3.81 \\
    OLMO2 (13B)  & 21.82 & 4.58 \\
    \midrule
    DINOv2 (ViT-B)   & 7.26  & 1.40 \\
    DINOv2 (ViT-L)  & 8.90  & 1.72 \\
    \bottomrule
\end{tabular}
\vspace{+1em}
\caption{JAM framework analysis attached to pretrained backbones}
\label{tab:jam_scale_analysis}
\end{table}

\FloatBarrier

\section{Further Results}

\begin{table*}[h!]
\centering
\resizebox{1.\textwidth}{!}{
    \begin{tabular}{l|l|c||cccccc}
    \toprule
    {Language Backbone } & {Vision Backbone } & {Alignment Method} & \multicolumn{2}{c}{Replace Task} & \multicolumn{2}{c}{Add Task} & \multicolumn{2}{c}{Swap Task} \\
    \cmidrule[0.5pt](rl){4-9}
        {(Model Size)} & {(Model Size)} & {for JAM} & {Recall@1 (binary)} & {Recall@1 (5-way)} & {Recall@1 (binary)} & {Recall@1 (5-way)} & {Recall@1 (binary)} & {Recall@1 (5-way)} \\
    \midrule
    Gemma2 (2B) & MAE (SSL) & Con & 58.55 & 50.33 & 56.37 & 50.46 & 63.73 & 49.29 \\
     & & NegCon & 77.62 & 69.38 & 89.15 & 80.16 & 70.81 & 58.63 \\
     & & Spread & 89.09 & 71.16 & 94.35 & 84.98 & 79.05 & 58.63 \\
    \midrule
    Gemma2 (2B) & Swav (SSL) & Con & 68.08 & 62.34 & 65.25 & 54.41 & 62.60 & 59.21 \\
     & & NegCon & 83.31 & 76.53 & 95.20 & 91.63 & 80.75 & 67.04 \\
     & & Spread & 88.88 & 76.98 & 95.83 & 90.64 & 81.01 & 67.70 \\
    \midrule 
    Gemma2 (2B) & Swin (Sup) & Con & 64.94 & 58.17 & 56.87 & 50.66 & 62.91 & 59.52 \\
     & & NegCon & 85.07 & 78.35 & 96.39 & 91.66 & 77.93 & 70.92 \\
     & & Spread & 85.16 & 79.01 & 97.87 & 92.22 & 80.32 & 71.46 \\
    \midrule
    Gemma2 (2B) & ViT (Sup) & Con & 60.78 & 54.74 & 60.79 & 55.05 & 59.37 & 52.28 \\
     & & NegCon & 86.06 & 75.89 & 96.31 & 92.50 & 59.21 & 49.58 \\
     & & Spread & 86.35 & 75.26 & 95.37 & 89.62 & 82.31 & 67.72 \\
    \midrule
    Gemma2 (9B) & DINOv2 (ViT-L; 300M) & Con & 68.77 & 63.95 & 59.31 & 54.58 & 60.92 & 51.85 \\
     & & NegCon & 87.52 & 84.33 & 98.89 & 93.14 & 84.44 & 69.69 \\
     & & Spread & 89.66 & 83.34 & 98.89 & 96.30 & 84.44 & 73.38 \\
    \midrule
    Llama3.2 (1B) & MAE (SSL) & Con & 67.82 & 54.48 & 61.08 & 55.33 & 61.19 & 54.68 \\
     & & NegCon & 82.92 & 74.09 & 94.16 & 84.80 & 76.77 & 68.26 \\
     & & Spread & 86.23 & 74.96 & 96.85 & 92.03 & 80.47 & 68.67 \\
    \midrule
    Llama3.2 (1B) & Swav (SSL) & Con & 65.94 & 58.97 & 68.12 & 62.93 & 75.39 & 68.03 \\
     & & NegCon & 92.37 & 83.91 & 92.12 & 87.48 & 78.62 & 71.84 \\
     & & Spread & 88.63 & 79.24 & \colorbox{Salmon}{98.94} & \colorbox{Apricot}{94.75} & \colorbox{Salmon}{83.29} & \colorbox{Apricot}{69.99} \\
    \midrule
    Llama3.2 (1B) & Swin (Sup) & Con & 70.59 & 63.60 & 69.70 & 63.48 & 75.39 & 68.03 \\
     & & NegCon & 86.01 & 83.05 & 93.89 & 88.39 & 69.11 & 63.45 \\
     & & Spread & 87.24 & 76.69 & 94.72 & 94.16 & 77.62 & 68.84 \\
    \midrule
    Llama3.2 (1B) & ViT (Sup) & Con & 66.81 & 57.49 & 64.50 & 60.70 & 64.87 & 59.79 \\
     & & NegCon & 82.81 & 77.15 & 93.16 & 90.57 & 74.50 & 63.18 \\
     & & Spread & 87.56 & 76.13 & 97.38 & 92.23 & 74.23 & 57.79 \\
    \midrule
    Llama3.2 (3B) & DINOv2 (ViT-L) & Con & 65.24 & 59.14 & 62.38 & 57.9 & 73.12 & 65.18 \\
     & & NegCon & 84.54 & 82.66 & 92.61 & 88.16 & 75.93 & 70.27 \\
     & & Spread & 89.43 & 82.48 & 95.74 & 91.25 & 82.44 & 76.77 \\
    \midrule
    OLMo2 (7B) & MAE (SSL) & Con & 66.88 & 57.37 & 68.12 & 56.98 & 58.36 & 51.85 \\
     & & NegCon & 86.39 & 74.57 & 93.05 & 88.33 & 71.94 & 60.62 \\
     & & Spread & 83.05 & 77.84 & 96.39 & 90.64 & 79.32 & 64.52 \\
    \midrule
    OLMo2 (7B) & Swav (SSL) & Con & 66.38 & 61.12 & 67.38 & 62.58 & 63.75 & 58.65 \\
     & & NegCon & 88.18 & 83.31 & 92.12 & 88.50 & 82.23 & 69.98 \\
     & & Spread & \colorbox{Salmon}{93.27} & \colorbox{Apricot}{85.63} & 98.43 & 93.70 & 81.01 & 72.81 \\
    \midrule
    OLMo2 (7B) & Swin (Sup) & Con & 65.35 & 59.45 & 70.82 & 63.79 & 55.67 & 50.85 \\
     & & NegCon & 89.40 & 86.70 & 95.28 & 89.35 & 73.66 & 71.96 \\
     & & Spread & 91.32 & 82.11 & 97.87 & 94.72 & 80.16 & 70.81 \\
    \midrule
    OLMo2 (7B) & ViT (Sup) & Con & 70.56 & 64.79 & 69.98 & 58.86 & 69.15 & 60.36 \\
     & & NegCon & 84.61 & 80.77 & 94.16 & 87.86 & 81.74 & 67.30 \\
     & & Spread & 89.25 & 81.01 & 98.43 & 95.18 & 76.49 & 66.44 \\
    \midrule
    OLMo2 (13B) & DINOv2 (ViT-L) & Con & 65.86 & 68.68 & 59.21 & 62.52 & 62.28 & 52.12 \\
     & & NegCon & 89.36 & 82.68 & 93.7 & 91.66 & 78.89 & 67.57 \\
     & & Spread & 90.53 & 83.18 & 97.96 & 92.42 & 84.71 & 69.69 \\
    \bottomrule
    \end{tabular}
}
    \caption{Image-to-Text Retrieval Results of Joint Autoencoder Modulator (JAM) for Vision-Language Compositionality with wider set of pretrained backbones.}
    \label{tab:allresults}
\end{table*}

\begin{table*}[t!]
\centering
\resizebox{0.5\textwidth}{!}{
    \begin{tabular}{l|l||c}
    \toprule
    {Language Backbone } & {Vision Backbone } & {Winoground Text-score} \\
    \midrule
    Gemma2 (2B) & DINOv2 (ViT-B; 86M) & \textbf{58.75} \\
    & ResNet50 & 58.13 \\
    \midrule
    Llama3.2 (1B) & DINOv2 (ViT-B) & 57.5 \\
    & ResNet50 & 61.3 \\
    \midrule
     OLMo2 (7B) & DINOv2 (ViT-B) & 55 \\
     & ResNet50 & 57.5 \\
     \midrule
     \multicolumn{2}{c||}{Pretrained CLIP (ViT-B/32)} & 32.80 \\
    \bottomrule
    \end{tabular}
}
    \caption{Winoground Text-score results using JAM with Spread framework. Given the observation from Sugarcrepe in Table \ref{tab:spreadloss}, \ref{tab:furtherexp} that finetuning CLIP in small data regime leading to decrease in performance, we provide pretrained CLIP as a baseline. We also observe JAM with Spread method showing the best performance compared to the baseline.}
    \label{tab:winoground}
\end{table*}

\FloatBarrier
\section{Compute Systems/Resources}
All experiments were conducted in either set-ups: Apple Macbook Pro (M2 chip) or NVIDIA RTX 3080. We utilized NVIDIA RTX 3080 system for language models' feature extraction and M2 chip system for vision models' feature extraction. Running our JAM framework is done in both set-ups. Finetuning CLIP model is conducted in NVIDIA RTX 3080 system. 

\bibliography{references} 

@book{plato, 
    title={Republic (De Republica)}, 
    author={Plato}, 
    year={375 BC}
}

@article{platonic,
    title = {The Platonic Representation Hypothesis},
    author = {Minyoung Huh and Brian Cheung and Tongzhou Wang and Phillip Isola},
    journal = {Proceedings of the 41st International Conference of Machine Learning},
    year = {2024},
    DOI = {https://doi.org/10.48550/arXiv.2405.07987}
}

@inproceedings{wit,
    author = {Srinivasan, Krishna and Raman, Karthik and Chen, Jiecao and Bendersky, Michael and Najork, Marc},
    title = {WIT: Wikipedia-based Image Text Dataset for Multimodal Multilingual Machine Learning},
    year = {2021},
    isbn = {9781450380379},
    publisher = {Association for Computing Machinery},
    address = {New York, NY, USA},
    url = {https://doi.org/10.1145/3404835.3463257},
    doi = {10.1145/3404835.3463257},
    booktitle = {Proceedings of the 44th International ACM SIGIR Conference on Research and Development in Information Retrieval},
    pages = {2443–2449},
    numpages = {7},
    location = {Virtual Event, Canada},
    series = {SIGIR '21}
}

@article{stitching, 
    title = {Revisiting model stitching to compare neural representations}, 
    author = {Yamini Bansal and Preetum Nakkiran and Boaz Barak}, 
    journal = {Advances in neural information processing systems}, 
    year = {2021}
}

@misc{stitching2,
      title={Understanding image representations by measuring their equivariance and equivalence}, 
      author={Karel Lenc and Andrea Vedaldi},
      year={2015},
      eprint={1411.5908},
      archivePrefix={arXiv},
      primaryClass={cs.CV},
      url={https://arxiv.org/abs/1411.5908}, 
}

@article{cca, 
    title = {Relations Between Two Sets of Variates},
    author = {Harold Hotelling}, 
    booktitle = {Biometrika}, 
    volume = {28}, 
    number = {3-4},
    pages = {321–377}, 
    year = {1936}, 
    DOI = {https://doi.org/10.1093/biomet/28.3-4.321}
}

@article{cka, 
    title = {Similarity of neural network representations revisited}, 
    author = {Kornblith, S. and Norouzi, M. and Lee, H. and Hinton, G}, 
    journal = {Proceedings of the 36th International Conference on Machine Learning}, 
    pages = {3519–3529}, 
    year = {2019}
}

@article{nearest_neighbors,
    author = {Klabunde, Max and Schumacher, Tobias and Strohmaier, Markus and Lemmerich, Florian},
    title = {Similarity of Neural Network Models: A Survey of Functional and Representational Measures},
    year = {2025},
    issue_date = {September 2025},
    publisher = {Association for Computing Machinery},
    address = {New York, NY, USA},
    volume = {57},
    number = {9},
    issn = {0360-0300},
    url = {https://doi.org/10.1145/3728458},
    doi = {10.1145/3728458},
    journal = {ACM Comput. Surv.},
    month = may,
    articleno = {242},
    numpages = {52}
}

@article{svcca, 
    title = {SVCCA: Singular vector canonical correlation analysis for deep learning dynamics and interpretability}, 
    author = {Raghu, M. and  Gilmer, J. and Yosinski, J. and Sohl-Dickstein, J.}, 
    journal = {Advances in neural information processing systems}, 
    year = {2017}
}

@inproceedings{clip,
  title={Learning Transferable Visual Models From Natural Language Supervision},
  author={Alec Radford and Jong Wook Kim and Chris Hallacy and Aditya Ramesh and Gabriel Goh and Sandhini Agarwal and Girish Sastry and Amanda Askell and Pamela Mishkin and Jack Clark and Gretchen Krueger and Ilya Sutskever},
  booktitle={International Conference on Machine Learning},
  year={2021},
}

@article{zhai2023sigmoid,
  title = {Sigmoid loss for language image pre-training},
  author = {Zhai, Xiaohua and Mustafa, Basil and Kolesnikov, Alexander and Beyer, Lucas},
  journal = {arXiv preprint arXiv:2303.15343},
  year = {2023}
}

@inproceedings{
    yuksekgonul2023when,
    title={When and why Vision-Language Models behave like  Bags-of-Words, and what to do about it?},
    author={Mert Yuksekgonul and Federico Bianchi and Pratyusha   Kalluri and Dan Jurafsky and James Zou},
    booktitle={International Conference on Learning Representations},
    year={2023},
    url={https://openreview.net/forum?id=KRLUvxh8uaX}
}

@article{chen2022perfectly,
  author = {Mayee F. Chen and Daniel Y. Fu and Avanika Narayan and Michael Zhang and Zhao Song and Kayvon Fatahalian and Christopher R\'e},
  title = {Perfectly Balanced: Improving Transfer and Robustness of Supervised Contrastive Learning},
  booktitle = {Proceedings of the 39th International Conference on Machine Learning (ICML 2022)},
  year = {2022},
}

@inproceedings{fu2022details,
  author = {Daniel Y. Fu and Mayee F. Chen and Michael Zhang and Kayvon Fatahalian and Christopher R\'e},
  title = {The Details Matter: Preventing Class Collapse in Supervised Contrastive Learning},
  journal = {Workshop on Artificial Intelligence with Biased or Scarce Data (AIBSD) at the 36th AAAI Conference on Artificial Intelligence},
  year = {2022},
}

@inproceedings{hsieh2023sugarcrepe,
  title = {SugarCrepe: Fixing Hackable Benchmarks for Vision-Language Compositionality},
  author = {Hsieh, Cheng-Yu and Zhang, Jieyu and Ma, Zixian and Kembhavi, Aniruddha and Krishna, Ranjay},
  booktitle = {Thirty-Seventh Conference on Neural Information Processing Systems Datasets and Benchmarks Track},
  year = {2023}
}

@inproceedings{thrush_and_ross2022winoground,
  author = {Tristan Thrush and Ryan Jiang and Max Bartolo and Amanpreet Singh and Adina Williams and Douwe Kiela and Candace Ross},
  title = {Winoground: Probing vision and language models for visio-linguistic compositionality},
  booktitle = {CVPR},
  year = 2022,
}

@inproceedings{cl,
    author = {Bengio, Yoshua and Louradour, J\'{e}r\^{o}me and Collobert, Ronan and Weston, Jason},
    title = {Curriculum learning},
    year = {2009},
    url = {https://doi.org/10.1145/1553374.1553380},
    doi = {10.1145/1553374.1553380},
    booktitle = {Proceedings of the 26th Annual International Conference on Machine Learning},
    pages = {41–48},
}

@inproceedings{
    adamw,
    title={Decoupled Weight Decay Regularization},
    author={Ilya Loshchilov and Frank Hutter},
    booktitle={International Conference on Learning Representations},
    year={2019},
    url={https://openreview.net/forum?id=Bkg6RiCqY7},
    }

@Article{supcon,
    title   = {Supervised Contrastive Learning},
    author  = {Prannay Khosla and Piotr Teterwak and Chen Wang and Aaron Sarna and Yonglong Tian and Phillip Isola and Aaron Maschinot and Ce Liu and Dilip Krishnan},
    journal = {arXiv preprint arXiv:2004.11362},
    year    = {2020},
}

@misc{vaswani2023attentionneed,
      title={Attention Is All You Need}, 
      author={Ashish Vaswani and Noam Shazeer and Niki Parmar and Jakob Uszkoreit and Llion Jones and Aidan N. Gomez and Lukasz Kaiser and Illia Polosukhin},
      year={2023},
      eprint={1706.03762},
      archivePrefix={arXiv},
      url={https://arxiv.org/abs/1706.03762}, 
}

@misc{ba2016layernormalization,
      title={Layer Normalization}, 
      author={Jimmy Lei Ba and Jamie Ryan Kiros and Geoffrey E. Hinton},
      year={2016},
      archivePrefix={arXiv},
      url={https://arxiv.org/abs/1607.06450}, 
}

@misc{shazeer2020gluvariantsimprovetransformer,
      title={GLU Variants Improve Transformer}, 
      author={Noam Shazeer},
      year={2020},
      archivePrefix={arXiv},
      url={https://arxiv.org/abs/2002.05202}, 
}

@article{regae,
    author = {Alain, Guillaume and Bengio, Yoshua},
    title = {What regularized auto-encoders learn from the data-generating distribution},
    year = {2014},
    issue_date = {January 2014},
    publisher = {JMLR.org},
    volume = {15},
    number = {1},
    issn = {1532-4435},
    journal = {J. Mach. Learn. Res.},
    month = jan,
    pages = {3563–3593},
    numpages = {31}
}

@inproceedings{regae_pc, 
    author = {Bao, Xuchan and Lucas, James and Sachdeva, Sushant and Grosse, Roger},
    title = {Regularized linear autoencoders recover the principal components, eventually},
    year = {2020},
    isbn = {9781713829546},
    booktitle = {Proceedings of the 34th International Conference on Neural Information Processing Systems},
    articleno = {585},
    numpages = {11},
    location = {Vancouver, BC, Canada},
    series = {NIPS '20}
}

@misc{kim2018interpretabilityfeatureattributionquantitative,
      title={Interpretability Beyond Feature Attribution: Quantitative Testing with Concept Activation Vectors (TCAV)}, 
      author={Been Kim and Martin Wattenberg and Justin Gilmer and Carrie Cai and James Wexler and Fernanda Viegas and Rory Sayres},
      year={2018},
      eprint={1711.11279},
      archivePrefix={arXiv},
      primaryClass={stat.ML},
      url={https://arxiv.org/abs/1711.11279}, 
}

@inproceedings{netdissect2017,
  title={Network Dissection: Quantifying Interpretability of Deep Visual Representations},
  author={Bau, David and Zhou, Bolei and Khosla, Aditya and Oliva, Aude and Torralba, Antonio},
  booktitle={Computer Vision and Pattern Recognition},
  year={2017}
}

@inproceedings{jia2021scaling,
  title={Scaling up visual and vision-language representation learning with noisy text supervision},
  author={Jia, Chao and Yang, Yinfei and Xia, Ye and Chen, Yi-Ting and Parekh, Zarana and Pham, Hieu and Le, Quoc and Sung, Yun-Hsuan and Li, Zhen and Duerig, Tom},
  booktitle={International conference on machine learning},
  pages={4904--4916},
  year={2021},
  organization={PMLR}
}

@inproceedings{li2022blip,
      title={BLIP: Bootstrapping Language-Image Pre-training for Unified Vision-Language Understanding and Generation}, 
      author={Junnan Li and Dongxu Li and Caiming Xiong and Steven Hoi},
      year={2022},
      booktitle={ICML},
}

@inproceedings{li2023blip2,
      title={{BLIP-2:} Bootstrapping Language-Image Pre-training with Frozen Image Encoders and Large Language Models}, 
      author={Junnan Li and Dongxu Li and Silvio Savarese and Steven Hoi},
      year={2023},
      booktitle={ICML},
}

@inproceedings{morcos2018insights,
  title={Insights on representational similarity in neural networks with canonical correlation},
  author={Morcos, Ari S and Raghu, Maithra and Bengio, Samy},
  booktitle={Advances in Neural Information Processing Systems},
  volume={31},
  year={2018}
}

@article{deepseekvl2024,
  title={DeepSeek-VL: Scaling Vision-Language with Decoupled Multimodal Pretraining},
  author={Haoyu Lu and Wen Liu and Bo Zhang and Bingxuan Wang and Kai Dong and Bo Liu and Jingxiang Sun and Tongzheng Ren and Zhuoshu Li and Hao Yang and Yaofeng Sun and Chengqi Deng and Hanwei Xu and Zhenda Xie and Chong Ruan},
  journal={arXiv preprint arXiv:2403.09696},
  year={2024}
}

@misc{openai2023gpt4v,
  title={GPT-4 with vision},
  author={OpenAI},
  year={2023},
  howpublished={\url{https://cdn.openai.com/papers/GPTV_System_Card.pdf}},
}

@misc{gemini2023,
  title={Gemini: a family of highly capable multimodal models},
  author={Google},
  year={2023},
  journal={arXiv preprint arXiv:2312.11805}
}

@article{gemma_2024,
    title={Gemma},
    url={https://www.kaggle.com/m/3301},
    DOI={10.34740/KAGGLE/M/3301},
    publisher={Kaggle},
    author={Gemma Team},
    year={2024}
}

@article{llama3modelcard,
    title={Llama 3 Model Card},
    author={AI@Meta},
    year={2024},
    url = {https://github.com/meta-llama/llama3/blob/main/MODEL_CARD.md}
}

@misc{oquab2023dinov2,
  title={DINOv2: Learning Robust Visual Features without Supervision},
  author={Oquab, Maxime and Darcet, Timothée and Moutakanni, Theo and Vo, Huy V. and Szafraniec, Marc and Khalidov, Vasil and Fernandez, Pierre and Haziza, Daniel and Massa, Francisco and El-Nouby, Alaaeldin and Howes, Russell and Huang, Po-Yao and Xu, Hu and Sharma, Vasu and Li, Shang-Wen and Galuba, Wojciech and Rabbat, Mike and Assran, Mido and Ballas, Nicolas and Synnaeve, Gabriel and Misra, Ishan and Jegou, Herve and Mairal, Julien and Labatut, Patrick and Joulin, Armand and Bojanowski, Piotr},
  journal={arXiv:2304.07193},
  year={2023}
}

@inproceedings{he2016deep,
  title={Deep residual learning for image recognition},
  author={He, Kaiming and Zhang, Xiangyu and Ren, Shaoqing and Sun, Jian},
  booktitle={Proceedings of the IEEE conference on computer vision and pattern recognition},
  pages={770--778},
  year={2016}
}

@Article{mae,
  author  = {Kaiming He and Xinlei Chen and Saining Xie and Yanghao Li and Piotr Doll{\'a}r and Ross Girshick},
  journal = {arXiv:2111.06377},
  title   = {Masked Autoencoders Are Scalable Vision Learners},
  year    = {2021},
}

@article{swav,
  title={Unsupervised Learning of Visual Features by Contrasting Cluster Assignments},
  author={Caron, Mathilde and Misra, Ishan and Mairal, Julien and Goyal, Priya and Bojanowski, Piotr and Joulin, Armand},
  booktitle={Proceedings of Advances in Neural Information Processing Systems (NeurIPS)},
  year={2020}
}

@inproceedings{swin,
  title={Swin Transformer: Hierarchical Vision Transformer using Shifted Windows},
  author={Liu, Ze and Lin, Yutong and Cao, Yue and Hu, Han and Wei, Yixuan and Zhang, Zheng and Lin, Stephen and Guo, Baining},
  booktitle={Proceedings of the IEEE/CVF International Conference on Computer Vision (ICCV)},
  year={2021}
}

@article{dosovitskiy2020vit,
  title={An Image is Worth 16x16 Words: Transformers for Image Recognition at Scale},
  author={Dosovitskiy, Alexey and Beyer, Lucas and Kolesnikov, Alexander and Weissenborn, Dirk and Zhai, Xiaohua and Unterthiner, Thomas and  Dehghani, Mostafa and Minderer, Matthias and Heigold, Georg and Gelly, Sylvain and Uszkoreit, Jakob and Houlsby, Neil},
  journal={ICLR},
  year={2021}
}

@inproceedings{openclip_laion,
  title={{LAION}-5B: An open large-scale dataset for training next generation image-text models},
  author={Christoph Schuhmann and
          Romain Beaumont and
          Richard Vencu and
          Cade W Gordon and
          Ross Wightman and
          Mehdi Cherti and
          Theo Coombes and
          Aarush Katta and
          Clayton Mullis and
          Mitchell Wortsman and
          Patrick Schramowski and
          Srivatsa R Kundurthy and
          Katherine Crowson and
          Ludwig Schmidt and
          Robert Kaczmarczyk and
          Jenia Jitsev},
  booktitle={Thirty-sixth Conference on Neural Information Processing Systems Datasets and Benchmarks Track},
  year={2022},
  url={https://openreview.net/forum?id=M3Y74vmsMcY}
}

@software{openclip,
  author = {Ilharco, Gabriel and
                  Wortsman, Mitchell and
                  Wightman, Ross and
                  Gordon, Cade and
                  Carlini, Nicholas and
                  Taori, Rohan and
                  Dave, Achal and
                  Shankar, Vaishaal and
                  Namkoong, Hongseok and
                  Miller, John and
                  Hajishirzi, Hannaneh and
                  Farhadi, Ali and
                  Schmidt, Ludwig},
  title = {OpenCLIP},
  month = jul,
  year = 2021,
  publisher = {Zenodo},
  version = {0.1},
  doi = {10.5281/zenodo.5143773},
  url = {https://doi.org/10.5281/zenodo.5143773}
}

@misc{olmo20242olmo2furious,
      title={2 OLMo 2 Furious}, 
      author={Team OLMo},
      year={2024},
      eprint={2501.00656},
      archivePrefix={arXiv},
      primaryClass={cs.CL},
      url={https://arxiv.org/abs/2501.00656}, 
}

@inproceedings{liang2024foundation,
  title={Foundation models for time series analysis: A tutorial and survey},
  author={Liang, Yuxuan and Wen, Haomin and Nie, Yuqi and Jiang, Yushan and Jin, Ming and Song, Dongjin and Pan, Shirui and Wen, Qingsong},
  booktitle={Proceedings of the 30th ACM SIGKDD conference on knowledge discovery and data mining},
  pages={6555--6565},
  year={2024}
}

@article{goswami2024moment,
  title={Moment: A family of open time-series foundation models},
  author={Goswami, Mononito and Szafer, Konrad and Choudhry, Arjun and Cai, Yifu and Li, Shuo and Dubrawski, Artur},
  journal={arXiv preprint arXiv:2402.03885},
  year={2024}
}

@article{talukdertotem,
  title={TOTEM: TOkenized Time Series EMbeddings for General Time Series Analysis},
  author={Talukder, Sabera J and Yue, Yisong and Gkioxari, Georgia},
  journal={Transactions on Machine Learning Research}
}

@inproceedings{das2024decoder,
  title={A decoder-only foundation model for time-series forecasting},
  author={Das, Abhimanyu and Kong, Weihao and Sen, Rajat and Zhou, Yichen},
  booktitle={Forty-first International Conference on Machine Learning},
  year={2024}
}

@inproceedings{hsic,
author = {Gretton, Arthur and Fukumizu, Kenji and Teo, Choon Hui and Song, Le and Sch\"{o}lkopf, Bernhard and Smola, Alexander J.},
title = {A kernel statistical test of independence},
year = {2007},
booktitle = {Proceedings of the 21st International Conference on Neural Information Processing Systems},
pages = {585–592},
numpages = {8},
location = {Vancouver, British Columbia, Canada},
series = {NIPS'07}
}

@misc{rkhs,
      title={Reproducing Kernel Hilbert Space, Mercer's Theorem, Eigenfunctions, Nystr\"om Method, and Use of Kernels in Machine Learning: Tutorial and Survey}, 
      author={Benyamin Ghojogh and Ali Ghodsi and Fakhri Karray and Mark Crowley},
      year={2021},
      eprint={2106.08443},
      archivePrefix={arXiv},
      primaryClass={stat.ML},
      url={https://arxiv.org/abs/2106.08443}, 
}

@misc{highdcca,
      title={High-Dimensional Canonical Correlation Analysis}, 
      author={Anna Bykhovskaya and Vadim Gorin},
      year={2025},
      eprint={2306.16393},
      archivePrefix={arXiv},
      primaryClass={econ.EM},
      url={https://arxiv.org/abs/2306.16393}, 
}

@misc{repcollapse,
      title={Understanding Dimensional Collapse in Contrastive Self-supervised Learning}, 
      author={Li Jing and Pascal Vincent and Yann LeCun and Yuandong Tian},
      year={2022},
      eprint={2110.09348},
      archivePrefix={arXiv},
      primaryClass={cs.CV},
      url={https://arxiv.org/abs/2110.09348}, 
}

\end{document}